\documentclass{article}

\usepackage[final]{neurips_2024}

\usepackage[utf8]{inputenc} 
\usepackage[T1]{fontenc}    
\usepackage{hyperref}       
\usepackage{url}            
\usepackage{booktabs}       
\usepackage{amsfonts}       
\usepackage{nicefrac}       
\usepackage{microtype}      
\usepackage{xcolor}         

\usepackage[utf8]{inputenc} 
\usepackage[T1]{fontenc}    
\usepackage{hyperref}       
\usepackage{url}            
\usepackage{booktabs}       
\usepackage{amsfonts}       
\usepackage{nicefrac}       
\usepackage{microtype}      
\usepackage{xcolor}         

\usepackage{amssymb} 
\usepackage{amsmath} 
\usepackage{makecell} 

\usepackage{graphicx}
\usepackage{pdfpages}
\usepackage{url}
\usepackage{xcolor}
\usepackage{epsfig}
\usepackage{adjustbox}
\usepackage{amsfonts}
\usepackage{amsmath}
\usepackage{amssymb}
\usepackage{booktabs} 
\usepackage{comment}
\usepackage{multirow}
\usepackage{caption}
\usepackage{subcaption}
\usepackage{textcomp}
\usepackage{relsize}
\usepackage{stmaryrd}
\usepackage{rotating}
\usepackage{helvet}
\usepackage{courier}
\usepackage{natbib}
\usepackage{lipsum}
\usepackage{tcolorbox} 
\usepackage{hyperref}
\usepackage{bbm}
\usepackage{algorithm}
\usepackage{algpseudocode}
\usepackage{colortbl} 
\usepackage{arydshln} 
\usepackage{tablefootnote}
\usepackage{mathrsfs}
\usepackage{cleveref}
\makeatletter
\AtBeginDocument{%
  \renewcommand{\sectionautorefname}{\S\@gobble}

  \renewcommand{\subsectionautorefname}{\S\@gobble}  
  \renewcommand{\subsubsectionautorefname}{\S\@gobble}  
  
}
\makeatother

\title{Train-Attention: Meta-Learning Where to Focus in Continual Knowledge Learning}

\author{
Yeongbin Seo~~~~~~~~~~~~~~~~~~
Dongha Lee \thanks{$^\ast$Co-corresponding authors}~~~~~~~~~~~~~~~~~~
Jinyoung Yeo $^\ast$\\
Department of Artificial Intelligence \\
Yonsei University\\
\texttt{\{suhcrates,donalee,jinyeo\}@yonsei.ac.kr}\\    
}

\begin{document}

\maketitle

\begin{abstract}
Previous studies on continual knowledge learning (CKL) in large language models (LLMs) have predominantly focused on approaches such as regularization, architectural modifications, and rehearsal techniques to mitigate catastrophic forgetting. However, these methods naively inherit the inefficiencies of standard training procedures, indiscriminately applying uniform weight across all tokens, which can lead to unnecessary parameter updates and increased forgetting. To address these shortcomings, we propose a novel CKL approach termed Train-Attention-Augmented Language Model (TAALM), which enhances learning efficiency by dynamically predicting and applying weights to tokens based on their usefulness. This method employs a meta-learning framework that optimizes token importance predictions, facilitating targeted knowledge updates and minimizing forgetting. Also, we observe that existing benchmarks do not clearly exhibit the trade-off between learning and retaining, therefore we propose a new benchmark, \textsc{LAMA-ckl}, to address this issue. Through experiments conducted on both newly introduced and established CKL benchmarks, TAALM proves the state-of-the-art performance upon the baselines, and also shows synergistic compatibility when integrated with previous CKL approaches. 
The code and the dataset will be available online\footnote{\url{https://github.com/ybseo-ac/TAALM}}

\end{abstract}

\begin{figure}[h!]
\centering
  \includegraphics[width=0.98\textwidth]{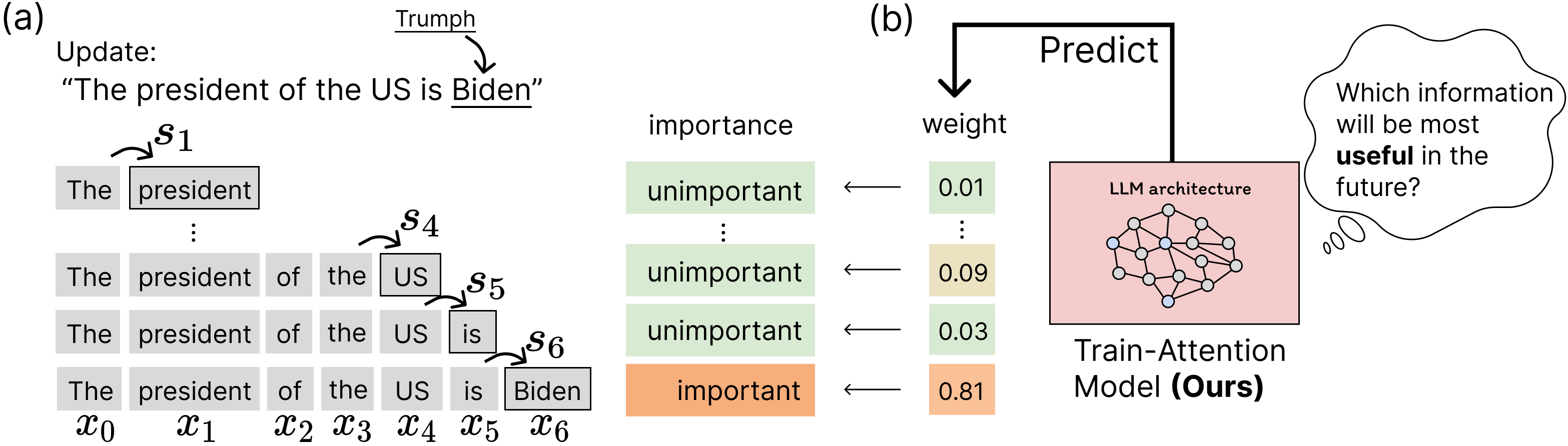}
  \caption{(a) Learning of Causal LM: The document is decomposed into multiple token sequences $s_i\doteq x_i|x_{<i}$\protect\footnotemark, which aligns with different importance, but uniformly weighted. (b) Train-Attention: Our proposed Train-Attention learns to predict weights that approximate importance, to enable targeted continual knowledge updates through label-free meta-learning method.}
  \label{fig:theme1}
\end{figure}

\footnotetext{In this work, we use the notation $x_i|x_{<i}$ to denote that the element $x_i$ is sequenced after the preceding elements $x_0,x_1,...,x_{i-1}$.}

\section{Introduction}

Large language models (LLMs), pre-trained on extensive text corpora, have demonstrated remarkable effectiveness when fine-tuned or prompted to perform a variety of downstream tasks \citep{brown2020language, raffel2020exploring, sanh2021multitask, wei2021finetuned}. However, as the world changes and new knowledge needs to be updated to the parameters, these models often suffer from a significant loss of previously learned knowledge (i.e., catastrophic forgetting \citep{kemker2018measuring, kirkpatrick2017overcoming}). To address this issue, the field of continual knowledge learning (CKL) is being actively researched \citep{jang2021towards, jang2022temporalwiki}, which aims to teach a model new knowledge while minimizing forgetting of previous knowledge. Previously explored approaches are broadly categorized into three: (1) minimizing parameter changes through regularization, (2) training the expanded parameters of the adapter while freezing the base model parameters, and (3) reviewing old knowledge. However, these approaches naively inherit the inefficiency of the standard fine-tuning procedure of causal LMs, which uniformly apply weights to all tokens, regardless of their importance.

This inefficiency of uniform weighting becomes more significant within the context of CKL, where the model is assumed to possess a substantial amount of world knowledge and grammatical capabilities already, thus emphasizing the need for limiting targets of learning. For example, consider a causal LM that has undergone both pre-training and fine-tuning and now requires to update the new information that ``The president of the US is Biden.'' Figure~\ref{fig:theme1}a illustrates how the model processes the example sentence. The only sequence that carries the essential information of this sentence is the final sequence $s_6$ (``The president of the US is'' $\rightarrow$ ``Biden''), which encapsulates the context of ``US'', ``president'', and ``Biden''. Conversely, another sequence such as $s_4$ (``The president of the'' $\rightarrow$ ``US'') only contains information that is already familiar to the model: the close association between ``president'' and the name of a nation, as well as the grammatical rule that a noun follows ``the''. Moreover, $s_1$ (``The'' $\rightarrow$ ``president'') introduces a harmful bias, suggesting that ``president'' should invariably follow ``The'', although any nouns could follow ``The''. If the model overemphasizes the likelihood of this sequence, several issues can arise: (1) Parameters will be updated more than the necessary amount to learn only essential information, thus resulting in more forgetting. (2) The training steps required to learn the important sequence could become prolonged.

Therefore, we hypothesize that focusing learning efforts on important tokens elevates the performance of the CKL. We present empirical evidence of this in \autoref{oracle-compare} (paragraph of the analysis on Oracle).
The concept of selecting important tokens has been previously explored outside the domain of CKL by \citet{hou2022token, lin2024rho}, and demonstrates enhanced performance on downstream tasks. These methods share the same principle, assigning more importance (we denote this ``token importance'') to the token with higher classification error, which assumes a definition of token importance as ``tokens with low-confidence are important''. While this approach can accelerate learning of low-confidence tokens, it is still not guaranteed that such low-confidence tokens are truly ``important''. This emphasizes a need for a more comprehensive definition of ``token importance''. To clarify this, in the example of \autoref{fig:theme1}, it is necessary to consider why human intuition easily accepts that the sequence $s_6$ is more important than others. This understanding comes from the anticipation that knowing the new president will be useful in the future (e.g., conversation with neighbors, school exams, etc) \citep{land1997knowledge}. Building on this concept, we define ``token importance'' as the expected utility of the token in related tasks, a concept we refer to as \textbf{usefulness}. Upon this definition of token importance, we propose a novel approach to CKL, named \textbf{T}rain-\textbf{A}ttention-\textbf{A}ugmented \textbf{L}anguage \textbf{M}odel (\textbf{TAALM}), which predicts weights of each token based on their usefulness, leveraging this weight on the training phase to enable efficient update of new knowledge. Train-Attention, the supportive model that predicts weight for the base model, is trained through the meta-learning method.

We also introduce a new CKL benchmark, \textbf{\textsc{LAMA-ckl}}, designed to offer a more clear comparison of learning and retention performance. This benchmark's advantages over the previous standard are explained in \autoref{sec:why_lama}. We experiment on \textsc{LAMA-ckl} and previous CKL benchmark (TemporalWiki \citep{jang2022temporalwiki}), and our method achieves remarkable \textbf{state-of-the-art} performance on both. Our method is compatible with other approaches, and shows enhanced performance when integrated, indicating a synergistic effect. We also compared RHO-1 \citep{lin2024rho}, which is the recent concurrent work on the token selecting method, where ours shows superior performance on CKL benchmarks. Our main contribution can be summarized in three. (1) We propose a novel token weighting approach to the CKL task, with a novel problem definition and meta-learning method. (2) A new benchmark for CKL based on the LAMA dataset. (3) Through extensive experiments, TAALM proves notable improvements over the baselines.

\section{Related Works}
\paragraph{Continual Knowledge Learning}

Continual Knowledge Learning (CKL) \citep{jang2021towards} is one variation of Continual Learning (CL), specified to LLM. It is more focused on updating new knowledge without catastrophic forgetting \citep{kirkpatrick2017overcoming, kemker2018measuring} of previously learned and preservable knowledge. 
Previous approaches for CL and CKL can be mainly categorized into three: regularization, architectural, and rehearsal. We analyze that the three approaches share a common goal; to minimize changes in parameters from initial points. \textbf{(1) Regularization}: directly controlling the extent of change in the parameters through weight regulation such as L2 \citep{kirkpatrick2017overcoming, zenke2017continual, lopez2017gradient, aljundi2018memory, chen2020recall}. \textbf{(2) Architectural}: freezing the base model parameters and expanding learnable parameters with adapters such as Lora \citep{houlsby2019parameter, hu2021lora, wang2020k, dettmers2024qlora}, thereby keeping initial parameters untouched. \textbf{(3) Rehearsal}: method of continually reviewing the data that is employed to train the initial model, ultimately returning the parameters to the initial points \citep{ shin2017continual, sun2019lamol, he2019mix, rolnick2019experience}. In this view, our method is another approach to achieve the same goal, minimizing change of parameters, by filtering objective tokens.

\paragraph{Meta-Learning}
Meta-learning \citep{finn2017model, hospedales2021meta} is most commonly understood as learning-to-learn; the process of improving a learning episode, over multiple outer learning episodes. During meta-learning, an outer (i.e., meta) learner is fitted to improve the learning of the inner (i.e., base) model. The meta-learner could be an initial parameter of the base model \citep{finn2017model}, an optimizer of the base model \citep{andrychowicz2016learning}, or a hyper-parameter of the base model such as learning-rate \citep{li2017meta, franceschi2018bilevel}. In this view, our meta-learner (Train-Attention) is an LLM architectural model that predicts hyper-parameters of the base model, as the token weights serve as the hyper-parameters in the training objective.

\paragraph{Token Selecting}
Methods to enhance learning by selecting specific tokens have previously been explored through various approaches: Token-Dropping \citep{hou2022token}, Focal Loss \citep{lin2017focal}, and RHO-1 \citep{lin2024rho}. These methods share a common principle: assigning more importance to the token with higher classification error.

\section{Train-Attention-Augmented Language Model (TAALM)}

\subsection{Token Importance and Token-Weighted Learning (TWL)}
\begin{eqnarray}
&\text{PPL}_\theta = -\frac{1}{N}\sum_i \log p(x_i | x_{<i};\theta) \quad \label{eq:celoss1} \\  
&\quad\quad\quad\quad\quad\quad\quad\quad\quad\quad = -\frac{1}{\sum_i w_0}\sum_i \log p(x_i | x_{<i};\theta) \times w_0  \ \ \  (w_0=1) \label{eq:celoss2} \\
&\text{Token Weighted (\textsc{tw}) PPL}_\theta = -\frac{1}{\sum_i w_i}\sum_i \log p(x_i | x_{<i} ;\theta)  \times w_i \quad\quad\quad\quad\quad\quad \ \ \label{eq:tw-celoss} \end{eqnarray}
To learn a document data $\mathcal{D}=\{x_1,...,x_N\}$ which is defined as a series of tokens, a dominant causal language model (LM) ($\theta$) commonly employs perplexity (PPL) as the objective function, as formalized in Eq.\eqref{eq:celoss1}. This can be also interpreted in the form of Eq.\eqref{eq:celoss2}, which assigns a uniform weight ($w_0=1$) to log probabilities of each sequence $x_i|x_{<i}$ across all documents. In contrast, our proposed methodology assigns weights $0 < w_i \leq 1$ to log probabilities of each sequence, which approximates the importance of each sequence, named token importance \citep{hou2022token}. We denote this set of weights as the \textbf{token weight}, and the training process that incorporates the weights (Eq.\eqref{eq:tw-celoss}) as \textbf{Token Weighted Learning (TWL)}.

\subsection{Train-Attention: Meta-Learning to Predict Token Importance}

\begin{figure}[h!]
\centering
  \includegraphics[width=0.99\textwidth]{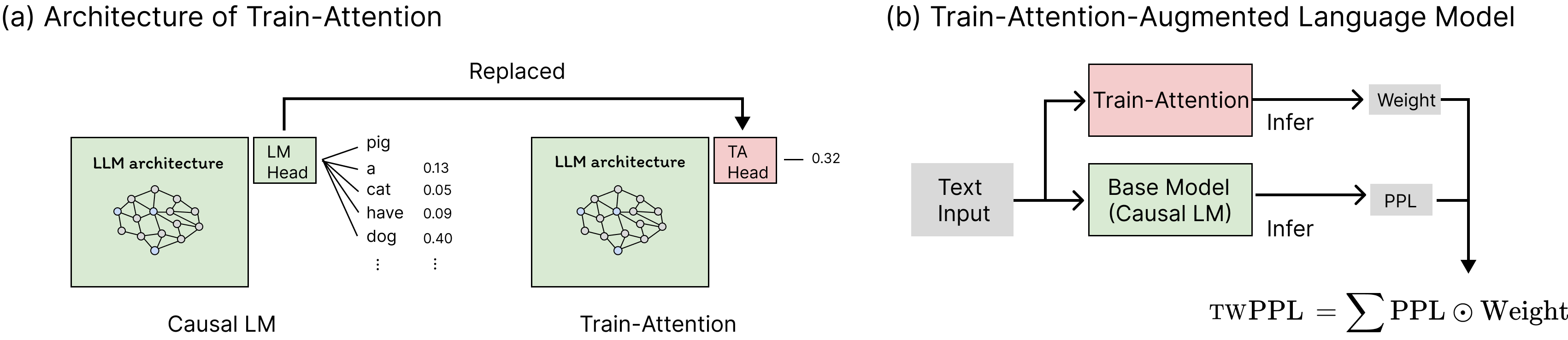}
  \caption{(a) depicts the architecture of Train-Attention, which shares the structure of causal LM, while the decoder layer (LM head) of causal LM is replaced from a linear layer of [$hidden\ size \times vocab\ size$] dimension to [$hidden\ size \times 1$] dimension, which is TA (Train-Attention) head. (b) depicts the TAALM, where the Train-Attention ($\phi$) is augmented to the base model ($\theta$).}
  \label{fig:architecture_inference}
\end{figure}

We suggest defining token importance as \textbf{usefulness}, which indicates how much the contained information is useful for solving related tasks in the future. Under this definition, a meta-learning approach can be derived to develop a supportive model (meta-learner) that predicts the optimal token weights. Let $\theta$ represent a base causal LM that continually learns knowledge and solves tasks. $\mathcal{T_D}$ represents a task that can be solved using information contained in $\mathcal{D}$. Training dataset $\mathscr{D}$ is a set of pairs of $\mathcal{D}$ and $\mathcal{T_D}$. We assume a task $\mathcal{T_D}$ can be defined as any type (e.g., predicting object labels, classification) as long as the performance can be measured in a differentiable form. The set of token weights, denoted as $W_\mathcal{D}$, comprises weights $w_i$ that represent the importance of each sequence $x_i|x_{<i}$ within $\mathcal{D}$. The meta-learner, named Train-Attention and denoted as $\phi$, predicts $W_\mathcal{D}$ from $\mathcal{D}$.
As illustrated in Figure~\ref{fig:architecture_inference}a, $\phi$ inherits the architecture and pretrained parameters of the causal LM, but the decoder layer is adjusted to yield only a single-dimensional float between [0,1] for each position. 

The desired process, learning knowledge and solving a task, is described in two steps; (a) \textit{learn}: $\theta$ is trained on $\mathcal{D}$ and is updated to $\theta'$. This update occurs in a TWL manner, with a token weight $W_{\mathcal{D},\phi}\leftarrow \phi(\mathcal{D})$ that $\phi$ predicts upon observing the data $\mathcal{D}$. (b) \textit{solve}: The revised model $\theta'$ is applied to solve the task $\mathcal{T_D}$, and the loss value $\mathcal{L}_{\theta'}(\mathcal{T_D})$ is computed to quantify the performance on $\mathcal{T_D}$, where $\mathcal{L}$ stands for loss function.

Two steps can be regarded as one black box function, which receives $\phi$ as an input and outputs $\mathcal{L}_{\theta'}(\mathcal{T_D})$. In other words, the task performance of $\theta'$ depends on how $\phi$ gives attention when learning evidence text data. And $\phi$ can be optimized to minimize the $\mathcal{L}_{\theta'}(\mathcal{T_D})$. For this, the procedure of (a) and (b) is developed to corresponding steps of Eq.\eqref{eq:phi5} and \eqref{eq:phi6}, where $\alpha$ and $\beta$ are the learning rates for each respective update.
\begin{eqnarray}
&\theta' \leftarrow \theta - \alpha \nabla_\theta \textsc{tw}\text{PPL}_\theta(\mathcal{D}, W_{\mathcal{D},\phi}) \label{eq:phi5} \\
&\phi \leftarrow \phi - \beta \nabla_\phi \mathcal{L}_{\theta'}(\mathcal{T_D})
\quad\quad\quad\quad \ \ \label{eq:phi6}\end{eqnarray}

\noindent
\begin{minipage}[b!]{\textwidth}
\begin{minipage}{.5\textwidth}
\hrule height 0.2pt
\captionof{algorithm}{Optimization of Train-Attention}\label{alg:cap}
\begin{algorithmic}[]
\hrule height 0.2pt
\Require Dataset $\mathscr{D}=\{(\mathcal{D}, \mathcal{T_D})\}$
\Require Learning rate for $\theta$, $\phi$ : $\alpha$, $\beta$
\Require Max iteration step for training $\theta$ : $M$
\State Initialize: base model ($\theta$), Train-Attention ($\phi$)
\While{ $\phi$ not converged}
    \State Sample a data pair ($\mathcal{D}, \mathcal{T_D}) \sim \mathscr{D}$
    \State $W_{\mathcal{D},\phi} \leftarrow \phi(\mathcal{D})$ \Comment{Predict weights}
    \For{$M$ times}
        \State $L_{learn} = \textsc{tw}\text{PPL}_{\theta}(\mathcal{D}, W_{\mathcal{D},\phi})$  
        \State Evaluate $\nabla_{\theta} L_{learn}$
        \State Update: $\theta \leftarrow \theta - \alpha \nabla_{\theta} L_{learn}$
    \EndFor
    \State $(\theta \text{ updated to } \theta')$
    \State $L_{solve} = \mathcal{L}_{\theta'}(\mathcal{T_D})$
    \State Evaluate $\nabla_{\phi} L_{solve}$
    \State Update: $\phi \leftarrow \phi - \beta \nabla_{\phi} L_{solve}$
    \State Reset $\theta'$ to initial point $\theta$
\EndWhile
\vspace{1mm}
\hrule height 0.2pt
\end{algorithmic}
\end{minipage}%
\begin{minipage}[]{.56\textwidth}
  \centering
  \includegraphics[width=0.7\textwidth]{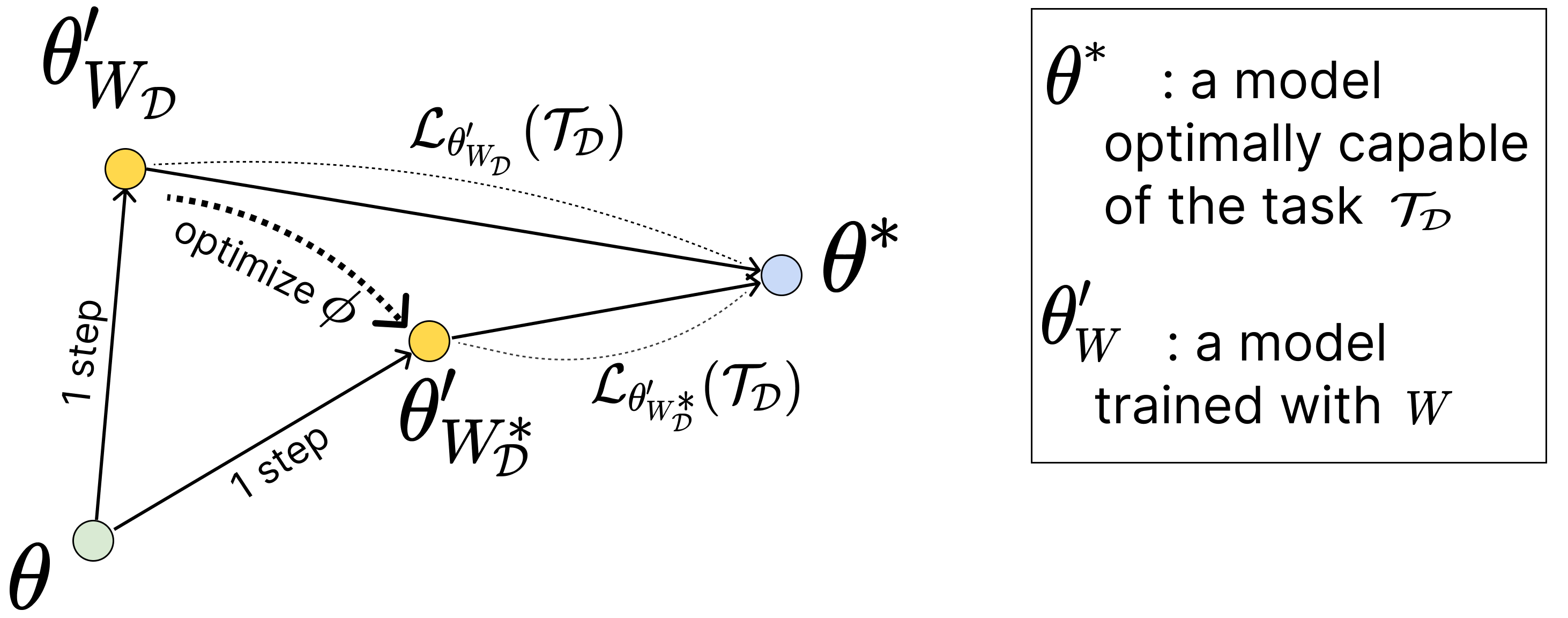}
  
  \captionof{figure}[]{Optimal $W$ leads $\theta$ closer to $\theta^*$.} \label{fig:metalearning}
  \includegraphics[width=0.75\textwidth]{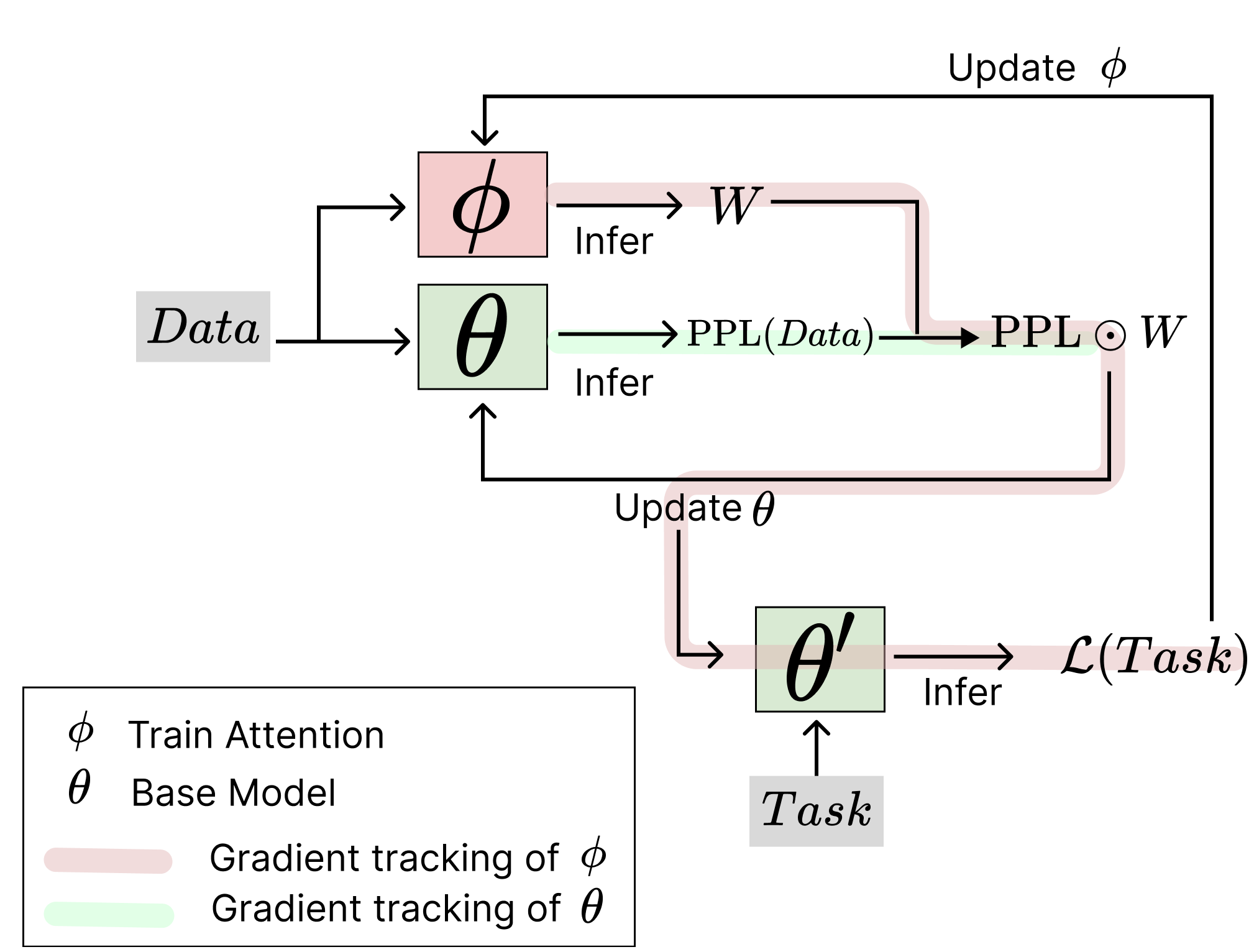}
  \captionof{figure}[]{One step update of $\phi$.} \label{fig:algorithm}
\end{minipage}
\end{minipage}

First, base model $\theta$ is updated to $\theta'$ through TWL, with the token weight $W_\mathcal{D}$ which is generated from $\phi$. Second, the meta-learner $\phi$ is updated based on the task performance, $\mathcal{L}_{\theta'}(\mathcal{T_D})$. These two steps of update can be also interpreted like Figure~\ref{fig:metalearning}. As the model $\theta$ steps out to a new state $\theta'$, the resulting position depends on which token weight ($W$) is applied. Some positions are closer to the $\theta^*$, a model optimally capable of the task $\mathcal{T_D}$, as the distance is measured with $\mathcal{L}_{\theta'}(\mathcal{T_D})$. We can conclude the weight with a shorter distance ($W_\mathcal{D}^*$) is more optimal, in the perspective of usefulness. To prevent the $\theta$ from converging to the point $\theta^*$, which disables the measurement of distances, we reset the model parameters to the initial state $\theta$ after every update of $\phi$. More detail is depicted in Figure~\ref{fig:algorithm} and Algorithm~\ref{alg:cap}. As the gradients of parameters of $\phi$ are tracked during the updating of $\theta$, its actual implementation is akin to the second derivative. The max iteration step of $\theta$ (denote as $M$ in Algorithm~\ref{alg:cap}) is fixed to 1 through our experiment. We employ gradient accumulation when updating $\phi$ for batch effect.

On the inference phase, $\theta$ learns data in TWL manner as in Eq.\eqref{eq:phi5}, with the parameter of $\phi$ frozen. This system is Train-Attention-Augmented Language Model (TAALM). In this work, the foundational structure of $\phi$ is fixed to the small model (TinyLlama-1.1B \citep{zhang2024tinyllama}), while it is augmented to both large (Llama2-7B \citep{touvron2023llama}) and small base models, because $\phi$ is compatible with any base model that shares the same tokenizer. Additionally, we explore utilizing a 101M-parameter bidirectional transformer (BERT) \citep{devlin2018bert} as a Train-Attention (TA) to further reduce resource requirements.

\section{Experiment}
We conduct experiments on two benchmarks. One is our newly designed \textsc{LAMA-ckl}, and the other is the established benchmark, \textsc{TemporalWiki} \citep{jang2022temporalwiki}.
We exclude the CKL benchmark by \citet{jang2021towards} which is not publicly available.
In this section, we present the corpus, evaluation setup, and training detail for Train-Attention and the test result within our proposed \textsc{LAMA-ckl} benchmark. For \textsc{TemporalWiki}, most configurations are aligned with the original work.

\subsection{\textsc{LAMA-ckl}}
\begin{figure}[h!]
\centering
  \includegraphics[width=1.01\textwidth]{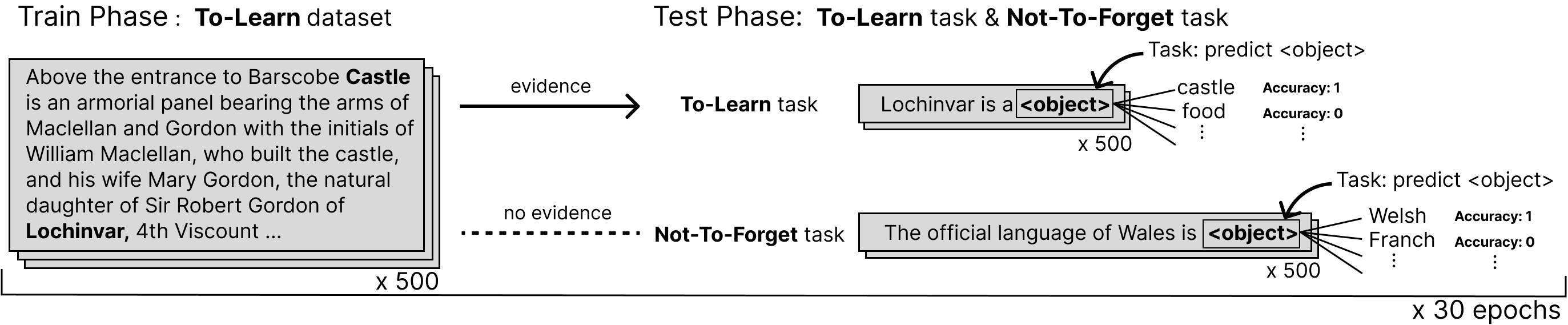}
  \caption{Evaluation procedure of the \textsc{LAMA-ckl} benchmark.}
  \label{fig:lama_ckl_illustration}
\end{figure}

For \textsc{LAMA-ckl}, we tailor the LAMA dataset (LAnguage Model Analysis \citep{petroni2019language}) to assess the CKL performance, especially the T-REx \citep{elsahar2018t} part which consists of data from Wikipedia and Wikidata. LAMA is a cluster of datasets that measures how much world knowledge is contained in the LLM. Each unit in the dataset includes a knowledge base triple \textsf{<subject, relation, object>}, along with corresponding evidence documents that support the information contained in this triple. Referring to the previous work \citep{jang2022temporalwiki}, a CKL benchmark should evaluate both \textbf{plasticity} and \textbf{stability}. Plasticity refers to how well the model updates new knowledge, while stability refers to how little the model forgets existing knowledge. Accordingly, we sample 500 units of \textsc{to-learn} and \textsc{not-to-forget} sets from LAMA to assess each dimension. During the evaluation, as illustrated on \autoref{fig:lama_ckl_illustration}, the model learns the evidence documents in the \textsc{to-learn} set. It is then tested on both the \textsc{to-learn} task and the \textsc{not-to-forget} task to assess plasticity and stability, respectively.

\paragraph{Dataset Setup}
Here, we outline a protocol for sampling test corpora for the \textsc{LAMA-ckl} benchmark. As \textsc{to-learn} set represents ``the knowledge that the model either encounters for the first time or needs to update'', it is selected based on two constraints: (1) sample from the categories of time-variant relations, predicated on the assumption that knowledge categorized as time-variant typically requires updates. (2) to ensure the concept of ``knowledge new to the model'', we select units where the task accuracy is zero when measured with pre-update baselines. Conversely, because \textsc{not-to-forget} set represents ``the knowledge that the model already knows and aims to retain'', it is selected from categories of time-invariant relations, with task accuracy of 1. We recommend sampling a new dataset by the specified constraints when evaluating models outside of the LLaMA-family, for more accurate assessment. The categorization of time-variant and time-invariant follows \citet{jang2021towards}. Each selected unit includes (1) an evidence document, (2) a knowledge base triple (e.g., \textsf{<Lochinvar, is an instance of, castle>}), and (3) a descriptive sentence encapsulating the triple (e.g., "Lochinvar is a castle"), which is inherited from LAMA dataset. The task is predicting object label tokens in the descriptive sentence. Details on data are in Appendix~\ref{app:lama-data-detail}

\paragraph{Evaluation Setup} \label{lama-ckl-eval-setup}
During the evaluation, each epoch consists of both a training phase and a test phase. In the training phase, the model is trained on a set of 500 evidence documents from the \textsc{to-learn} set. Subsequently, in the test phase, the model's prediction accuracy for the object labels is assessed using 500 descriptive sentences from both the \textsc{to-learn} and \textsc{not-to-forget} sets. This process is repeated over 30 epochs. For the \textsc{to-learn} set, an increase in mean accuracy from 0 signifies the model's plasticity. Conversely, a decline in mean accuracy for the \textsc{not-to-forget} set from 1 to lower values indicates the model's stability, as it tends to forget previously learned information.

In the proposed benchmark \textsc{LAMA-ckl}, we suggest four main factors as evaluation indicators. \textbf{1) Top Acc:} the highest \textsc{to-learn} accuracy among checkpoints of 30 epoch. \textbf{2) Epoch:} the epoch where the Top Acc appears. \textbf{3) NF Acc:} \textsc{not-to-forget} accuracy of the checkpoint model which is the same as Top Acc. \textbf{4) Total Knowledge}: the sum of Top Acc and NF Acc, indicating total capacity of knowledge including updating and maintaining. We chose these factors because the best CKL system is one that \textit{learns the most and the fastest and loses the least}. Factor 1, 3, and 4 are better if higher, while factor 2 is better if lower.
The detailed configurations for training datasets are in Appendix~\ref{app:lama-eval-detail}.

\begin{figure}[h!]
  \includegraphics[width=0.99\textwidth]{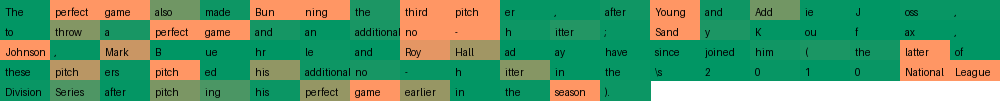}
  \caption{Heat map of token weights from Train-Attention. {\color{orange}Orange} color indicates higher weights.}
  \label{fig:lama-heatmap}
\end{figure}

\paragraph{Train-Attention Training Setup}
\label{lama-ta-train-setup}
Referring to Algorithm~\ref{alg:cap}, the training procedure of Train-Attention requires data $\mathcal{D}$ and related task $\mathcal{T_D}$. For \textsc{LAMA-ckl} dataset, we assign evidence document of each unit as $\mathcal{D}$, and knowledge base triple in a document of \textbf{schematic form} as $\mathcal{T_D}$. The perplexity of \textsf{object} token is assigned as the objective of $\phi$. \autoref{fig:lama-heatmap} shows the heat map of token-weight that Train-Attention generates. Train-Attention seems to give more attention to entities of certain categories, rather than words with general grammatical roles. We describe the detailed configuration and findings on the training of Train-Attention in Appendix~\ref{app:train-attention}.
\paragraph{Baseline Setup}
We utilize Llama2-7B integrated with QLoRA \citep{dettmers2024qlora} as a base model. The baseline methods and their hyper-parameter settings follow previous CKL study of \citet{jang2022temporalwiki}: standard finetune, K-Adapter \citep{wang2020k}, Mix-review \citep{he2019mix}, LoRA \citep{hu2021lora}, RecAdam \citep{chen2020recall}. We regard standard finetune on QLoRA as a substitute for full finetuning and LoRA, thus skipping the two baselines. We also compare RHO-1 \citep{lin2024rho}, which is the most recent concurrent work on the token selecting method, sharing a similar concept with ours. We chose the initial parameter state as a reference model, which is utilized to select important tokens for RHO-1, with other hyper-parameters following the optimal of the original. We also evaluate a model trained in TWL manner with \textbf{Oracle} token weight. For which, a weight of 1 is exclusively assigned to the \textsf{object} label token in the evidence document, and the rest is assigned zero weight. Oracle is compared for two purposes: \textbf{(1)} To prove the concept that token-weighted learning has the advantage for CKL. \textbf{(2)} To check the performance upper bound of Train-Attention. Detailed configurations are in Appendix~\ref{app:lama-baselines}

\subsubsection{Result \& Analysis}
\label{lama-ckl-result}

\begin{figure}[h!]
  \centerline{
 \begin{subfigure}{0.5\textwidth} 
    \centering
    \includegraphics[width=\linewidth]{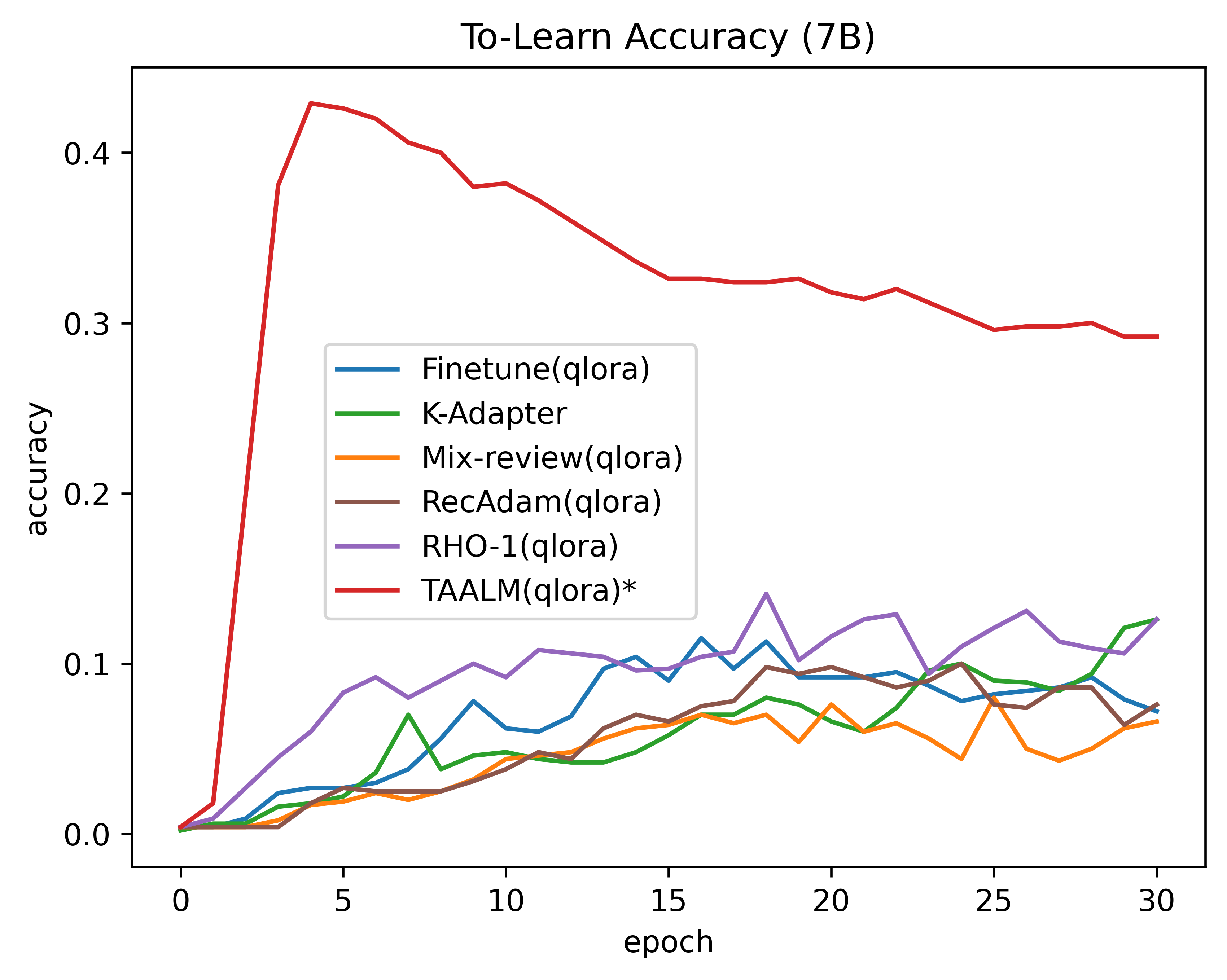}
    \end{subfigure}%
 \begin{subfigure}{0.5\textwidth} 
    \centering
    \includegraphics[width=\linewidth]{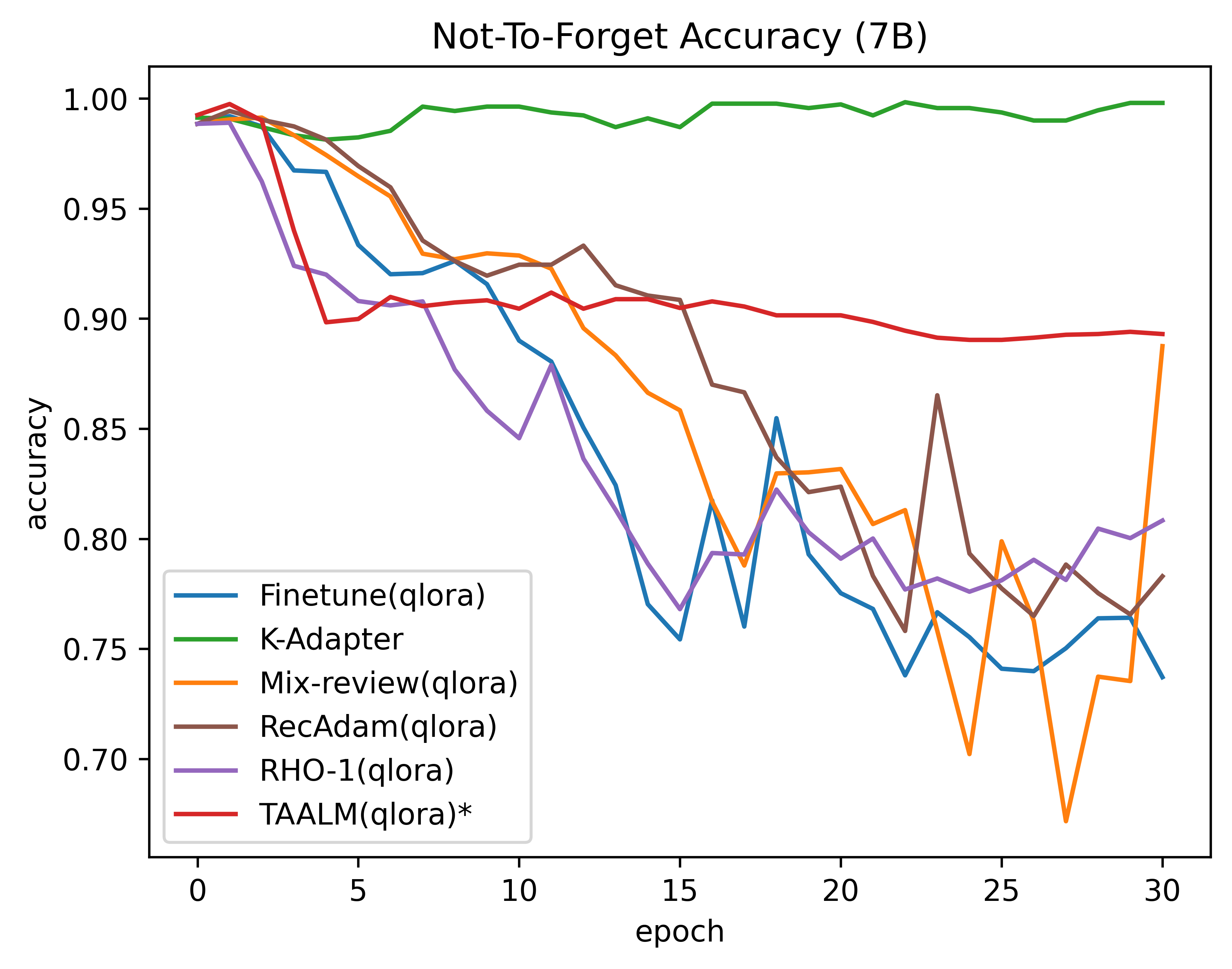}
    \end{subfigure}%
    }

  \caption{\textsc{LAMA-ckl} performance of large (Llama2-7B) baseline models. The graph on the left represents \textsc{to-learn} task, and the graph on the right represents \textsc{not-to-forget} task performance. The x-axis is the learning epoch, and the y-axis is accuracy.}
  
  \label{fig:lama_7b}
\end{figure}
\begin{table}[h!]
\caption{\textsc{LAMA-ckl} performance of Llama2-7B based baselines. The evaluation indicator of each column is explained on \autoref{lama-ckl-eval-setup}. The best performance is marked as \textbf{bold} while the second best is \underline{underlined}.}

\hfill

\label{tab:lama_7b}

\hfill

\centering
\begin{tabular}{
 lllll
}
\toprule
                          & Top Acc      & Epoch     & NF Acc &Total Knowledge   \\ \hline
Finetune(QLoRA)           & 0.1150        & \underline{16}     & 0.8174   & 0.9324 \\
K-Adapter                 &  0.1260     & 30 & \textbf{0.9980} & \underline{1.1240} \\
Mix-review(QLoRA)        &0.0800            &25   &0.7988 & 0.8788   \\
RecAdam(QLoRA)           &0.1000            &24   & 0.7933 &0.8933  \\
RHO-1(QLoRA)             & \underline{0.1410}      & 18  & 0.8223  & 0.9633\\  \hdashline \addlinespace
TAALM(QLoRA)           & \textbf{0.4290}      & \textbf{4}  & \underline{0.8983} & \textbf{1.3273}  \\
\bottomrule
\end{tabular}

\end{table}

\paragraph{TAALM achieves substantial CKL performance}
\label{sec:lama-result}
As results show in Figure~\ref{fig:lama_7b} and Table~\ref{tab:lama_7b}, our method (TAALM) overwhelms other baselines on ability and speed of learning. Ours records the Top Acc of \textsc{to-learn} task as 0.4290 on only 4 epochs of updating. The accuracy record of ours is 3.04 times higher than the second place (RHO-1). And the required epoch is only 25\% of the second place (standard finetune).

K-Adapter shows the highest NF Acc, rarely forgetting previous knowledge. However, the \textsc{to-learn} accuracy also barely increases, indicating that parameter updates rarely occur. This could be due to the architectural difference from QLoRA, the backbone of other baselines. Therefore, we conclude that it is meaningless to compare K-Adapter and QLoRA based baselines in the same condition. When comparing only the QLoRA base methods, TAALM shows overwhelming performance in all dimensions. Ours learn the most, the fastest, and forget the least. RHO-1, one of the token selecting methods, also shows a bigger capacity for learning and less forgetting than standard finetuning, However, the advantage is minimal compared to ours.

\paragraph{Combination with TAALM improves all of the previous baselines} 
As Train-Attention is an approach to manipulating loss values on the end side, it is easily compatible with previous methods based on other concepts. Thus we combine various methods to TAALM and observe their performance. Referring to the experimental results and details in Appendix~\ref{app:combination}, each combined version shows a highly improved capacity than the baseline alone. Especially, combining our method with K-Adapter demonstrates a considerable balance between stability and plasticity.

\paragraph{Oracle vs Train-Attention vs Standard finetune}
\phantomsection
\label{oracle-compare}

\begin{figure}[h!]
  \centerline{
 \begin{subfigure}{0.4\textwidth} 
    \centering
    \includegraphics[width=\linewidth]{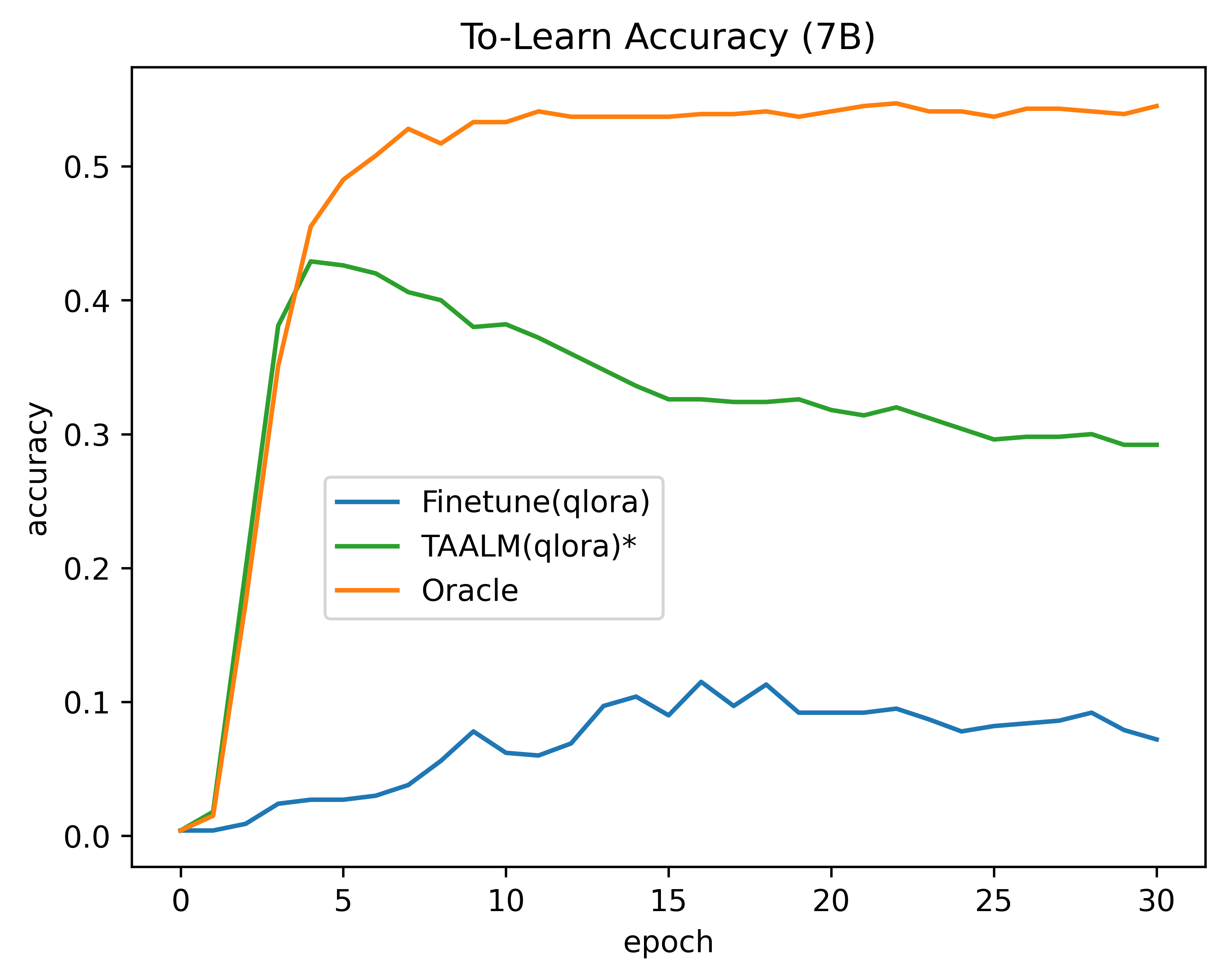}
    \end{subfigure}%
 \begin{subfigure}{0.4\textwidth} 
    \centering
    \includegraphics[width=\linewidth]{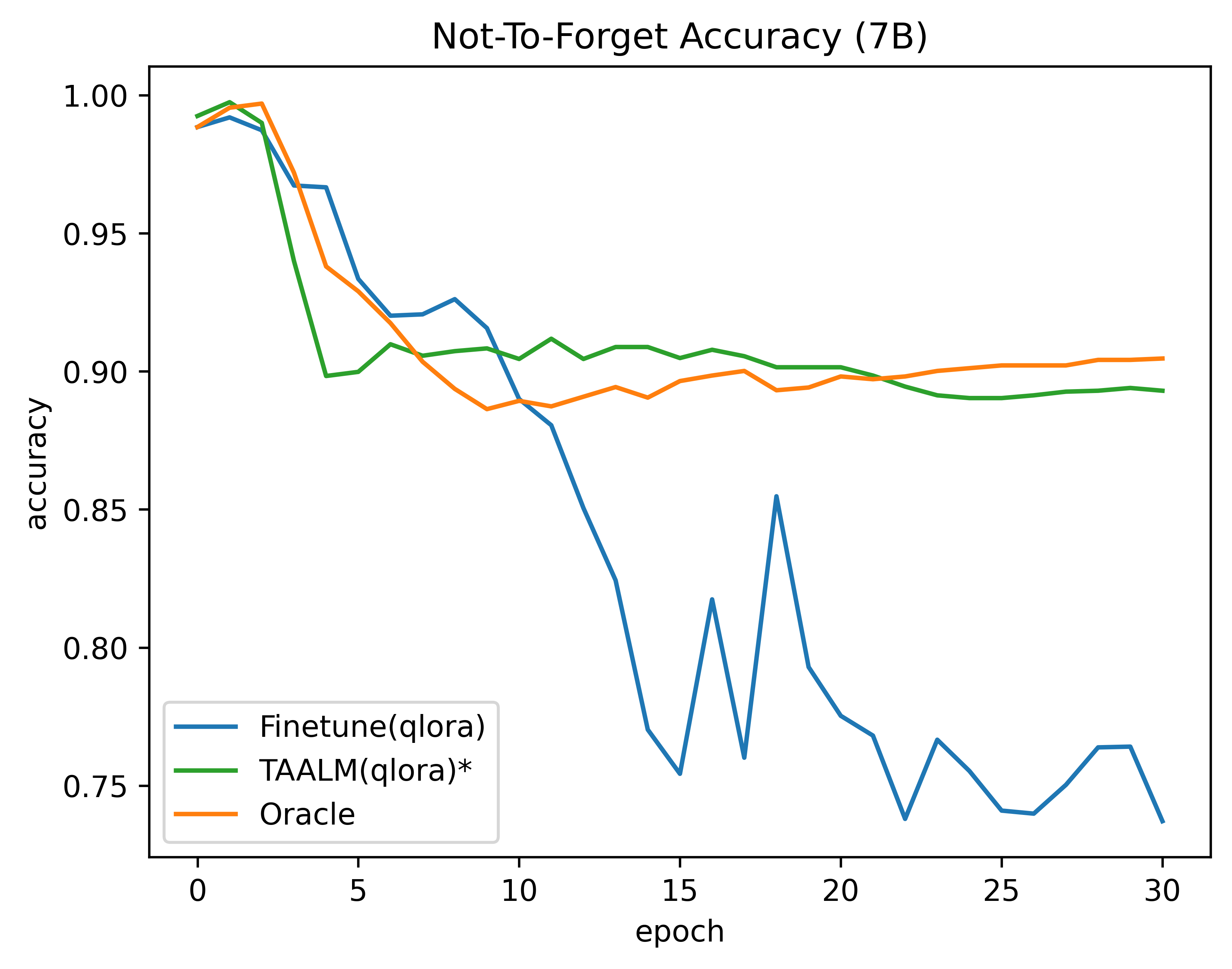}
    \end{subfigure}%
    }

  \caption{Comparison between Oracle, standard finetuning, and ours, tested on \textsc{LAMA-ckl}. }
  
  \label{fig:oracle}
\end{figure}

It is assumed that TWL with Oracle label is the upper bound of performance of TWL with Train-Attention, because it is sharply targeting only the necessary token. Referring to Figure~\ref{fig:oracle}, for Oracle, Top Acc is 0.5470, epoch 17, and NF Acc is 0.9002. Oracle shows that Top Acc is 4.75 times higher and NF Acc is 1.11 times higher than standard finetune, thus proving the substantial advantage of token-weighted learning on CKL. Also, TAALM nearly approaches Oracle, as it achieves 78.2\% of Oracle Top Acc. NF Acc of ours also maintains a similar level to Oracle, indicating that Train-Attention is optimized close to the upper bound. It also indicates that optimization through meta-learning could excel human labeled weight.

\paragraph{Small (1B) TAALM excels large baselines}
We also experiment on the smaller (TinyLlama-1B) baselines, and TAALM on 1B records the best compared to 7B baselines. This observation indicates that our method outperforms other baseline methods even with significantly smaller parameter sizes and computational resources. Related Table and Figure are on Appendix~\ref{app:lama-1b}.

\paragraph{TA on BERT achieves comparable performance with small resources}
We train BERT as a TA and evaluated it on the \textsc{LAMA-ckl} dataset. Training TA on BERT requires only a single 24GB GPU, significantly reducing resource usage compared to the previous model (single 82GB GPU), yet achieving performance similar to the larger (1.1B) TA (Appendix \ref{app:bert_taalm}).

\paragraph{Ablation study on various design choices} We conduct an ablation study on various design choices applied to the token importance predicted by TA: (1) masking out tokens (setting importance to 0) in real-time when the prediction matches the label, and (2) dropping weights with token importance below the top k\% threshold. The ablation study reveals that heuristic adjustments degrade performance, as TA is already in an optimized state. Details are in Appendix \ref{app:ablation}.

\subsection{\textsc{TemporalWiki}}
We experiment on the original CKL learning benchmark \textsc{TemporalWiki} \citep{jang2022temporalwiki}, where models have to continually learn Wikipedia documents of serial periods (\textsc{0809, 0910, 1011, 1112}) and test on the corresponding \textsc{Twiki-Probes}, which is a dataset of knowledge base triples. As we train Train-Attention on the \textsc{0809} data, tests are conducted on the rest. We experiment with only a small (TinyLlama-1B) model, which is bigger than the baselines of the original work (GPT-2 Large). We conduct a separate experiment on QLoRA-based K-Adapter based models, referring to the analysis on experiment of \textsc{LAMA-ckl} on \autoref{sec:lama-result}. We only consider training \textsc{Diffset}, which is the only changed part of Wikipedia, because it is reported as a condition of the best performance. Most of the experimental settings follow the original, and additional change is described in Appendix~\ref{app:temporalwiki}.

\begin{table*}[h!]
\caption{\textsc{TemporalWiki} performacne of small (TinyLlama-1B) baselines. \textbf{Un} refers \textsc{Unchanged}, \textbf{C} refers \textsc{Changed}, \textbf{Avg} refers the average of the two. TAALM is our method.} 
\label{twiki-result}
\begin{subtable}{\textwidth}
\subcaption{QLoRA based}
 \resizebox{\textwidth}{!}{\begin{tabular}{cccccccccccccc}
    \toprule
    \multicolumn{1}{l}{} & \multicolumn{3}{c}{TWiki-Probes-0910} & \multicolumn{3}{c}{TWiki-Probes-1011} &
    \multicolumn{3}{c}{TWiki-Probes-1112}
    \\ \cmidrule(lr){2-4} \cmidrule(lr){5-7} \cmidrule(lr){8-10}  &  \textbf{Un} & \textbf{C} & \textbf{Avg} & \textbf{Un} & \textbf{C} & \textbf{Avg} & \textbf{Un} & \textbf{C} & \textbf{Avg} \\
    \midrule
Finetune(QLoRA)           & 9.999    & 10.057 &10.028     &9.554      &9.531  & 9.543    & 9.736   &9.632  &  9.684   \\
Mix-review(QLoRA)         & 9.529      & 9.579  & 9.554    &9.514      & 9.486 & 9.501    & 9.562      & 9.452 & 9.507    \\
RecAdam(QLoRA)            & 9.514     & 9.604  & 9.559      &8.992      & 9.031  &9.012   & 9.579     & 9.479  &  9.529   \\ 
RHO-1(QLoRA)              & \underline{4.389}      & \underline{4.624} & \underline{4.507}    & \underline{4.360}     & \underline{4.395} & \underline{4.3775}    & \underline{4.471}     & \underline{4.717} &  \underline{4.594}   \\ \hdashline \addlinespace
TAALM(QLoRA) & \textbf{4.019}      & \textbf{4.268} & \textbf{4.1435}    & \textbf{4.030}& \textbf{4.154} & \textbf{4.092}    & \textbf{4.036} & \textbf{4.357} &  \textbf{4.197}   \\
\bottomrule
\end{tabular}}
\end{subtable}

\begin{subtable}{\textwidth}
\vspace{2mm}
\subcaption{K-Adapter based}
 \resizebox{\textwidth}{!}{\begin{tabular}{cccccccccccccc}
    \toprule
    \multicolumn{1}{l}{} & \multicolumn{3}{c}{TWiki-Probes-0910} & \multicolumn{3}{c}{TWiki-Probes-1011} &
    \multicolumn{3}{c}{TWiki-Probes-1112}
    \\ \cmidrule(lr){2-4} \cmidrule(lr){5-7} \cmidrule(lr){8-10}  &  \textbf{Un} & \textbf{C} & \textbf{Avg} & \textbf{Un} & \textbf{C} & \textbf{Avg} & \textbf{Un} & \textbf{C} & \textbf{Avg} \\
    \midrule
Finetune(K-Adapter)           & 2.768      &2.982 & 2.875  & 2.598     &2.679  & 2.639     & 2.552     & 2.669 &  2.611   \\
Mix-review(K-Adapter)         & 2.486      &2.724 & 2.605    & 2.334      & 2.446 & 2.390    & 2.284      & 2.425 & 2.355    \\
RecAdam(K-Adapter)         & 2.494      & 2.710 & 2.602  & 2.323      & 2.415 & 2.369    & 2.248      & 2.375 & 2.312    \\
RHO-1(K-Adapter)              & 2.722      & 2.950 & 2.836    & 2.538     &2.634 & 2.586    &  2.478     &  2.603 & 2.541    \\ \hdashline \addlinespace
TAALM (K-Adapter) on \textsc{LAMA-ckl} & \underline{2.364}  & \underline{2.601} & \underline{2.483}     & \underline{2.168}      & \underline{2.282} & \underline{2.225}    & \underline{2.123}      & \underline{2.307} & \underline{2.215}    \\
TAALM (K-Adapter) & \textbf{1.980}  & \textbf{2.194} & \textbf{1.9705}    & \textbf{1.907}      & \textbf{2.034} & \textbf{2.087}    & \textbf{1.901}      & \textbf{2.070} &  \textbf{1.986}   \\
\bottomrule

\end{tabular} }
\label{tab:twiki-kadapter}
\end{subtable}
\end{table*}

\subsubsection{Result \& Analysis}
\label{sec:twiki-result}

Referring to Table~\ref{twiki-result}, our method presents the state-of-the-art performance across both experiments on QLoRA based and K-Adapter based models. This achievement is consistent in all periods, and in both \textsc{Changed} and \textsc{Unchanged} \textsc{Twiki-Probes}. This result is aligns with the \textsc{LAMA-ckl} benchmark result, showing that our method has a substantial advantage on the CKL. QLoRA based baselines showed poor performance compared to K-Adapter based baselines, indicating architectural disadvantage. We also test TAALM optimized for \textsc{LAMA-ckl} on the \textsc{TemporalWiki}, referring Table~\ref{tab:twiki-kadapter}. It achieves the second-best performance, indicating robustness across different distributions of tasks.

\subsection{Why \textsc{LAMA-ckl}: clear contrast of plasticity and stability}
\label{sec:why_lama}
For the benchmark \textsc{TemporalWIki}, because \textsc{Diffset} is corpora of evidence documents for \textsc{Changed} set, learning of \textsc{Diffsets} is supposed to result in performance improvement over \textsc{Changed} set and forgetting of \textsc{Unchanged} set. However, during our experiments, we observe that both \textsc{Changed} and \textsc{Unchanged} performance tend to move in similar directions when learning \textsc{Diffset}, which is in contradiction to our assumption (Appendix~\ref{app:twiki-anal}). We analyze this for two primary reasons. First, the \textsc{Diffset} contains evidence documents for both the \textsc{Changed} and the \textsc{Unchanged} sets (Appendix~\ref{app:twiki-anal}). This is a complicating factor in the evaluation of stability. Second, the experimental setup involves training on a vast amount of data, an average of 707K documents per period, for just a single epoch at a low learning rate. This might result in learning little amount of knowledge, while the task ability is challenged by extensive iterations of updates; which is closer to a continual learning setup rather than CKL. To address this issue, we structured the \textsc{LAMA-ckl} as follows: (1) To mitigate the issue of data overlap, we partition the dataset into variant and invariant subsets. These subsets are further classified based on the task accuracy measured by pre-update baselines. (2) We conduct training over multiple epochs on a relatively small dataset to observe the acquirement of knowledge. In practice, our benchmark shows a clear upward trend in the \textsc{to-learn} set and a distinct decline in the \textsc{not-to-forget} set as training progresses, clearly demonstrating the contrast between plasticity and stability.

\section{Conclusion and Limitation}
\label{sec:conclusion}
In this paper, we demonstrate that the application of Train-Attention significantly enhances CKL performance and is also synergistic with other baselines. Nevertheless, our work has the following limitations and potential for future exploration.

\paragraph{Task specificity of Train-Attention }
Train-Attention is trained to focus on information related to tasks encountered in the training session. This allows task performance to increase, but on the other hand, it left questions as to whether it would be possible to cope with other tasks. Nonetheless, if the task entails the acquisition of general knowledge, it will be transferable to other tasks sharing similar distributions. For instance, TAALM optimized to \textsc{LAMA-ckl} also achieved the best performance on the \textsc{TemporalWiki} (\autoref{sec:twiki-result}). Additionally, Train-Attention can ever evolve to adapt, enabling optimal performance for the current tasks. 
\paragraph{What if there are no data-task pair}
Train-Attention can be trained only if there is a data-task pair. If there is no paired dataset, it can be doubted that training is difficult. However, every knowledge has its purpose, and we can find a strategy to discover it. \textbf{1) Search:} When the data and task pools are separate, we can join highly related pairs via searching. \textsc{TemporalWiki} is also a dataset in which data and tasks are not paired, thus we conduct a lexical search. In the future, also dense research methods can be explored. \textbf{2) Generate:} If there is even no separate task pool, we can at least get prior information about what kind of tasks are probable to come in the future. Synthetic tasks can be generated via instruction-tuned LLM, based on this prior information. These methods also resemble the human's cognitive strategy, who often revisit past memories and pose hypothetical questions to themselves to enhance the efficiency of their learning processes.

\paragraph{Broader impacts}
Our method aims to increase the ability of CKL, therefore enhancing the practicability of LLMs and saving the computational resources for fine-tuning entire huge LLMs. We believe that this paper does not have any immediate negative societal impact.

\section*{Acknowledgement}
This work was supported by STEAM R\&D Project, NRF, Korea (RS-2024-00454458) and Institute of Information \& Communications Technology Planning \& Evaluation (IITP) grant funded by the Korean government (MSIT)(No.RS-2020-II201361, Artificial Intelligence Graduate School Program (Yonsei University)) and (2022-0-00077, RS-2022-II220077,AI Technology Development for Commonsense Extraction, Reasoning, and Inference from Heterogeneous Data). Jinyoung Yeo and Dongha Lee are the co-corresponding authors.

\bibliography{main}
\bibliographystyle{abbrvnat}


\appendix

\newpage
\appendix
\section{\textsc{LAMA-ckl} Benchmark Additional Detail}

\subsection{Dataset setup detail}
\label{app:lama-data-detail}
\begin{figure}[h!]
  \makebox[\textwidth][c]{\includegraphics[width=0.99\textwidth]{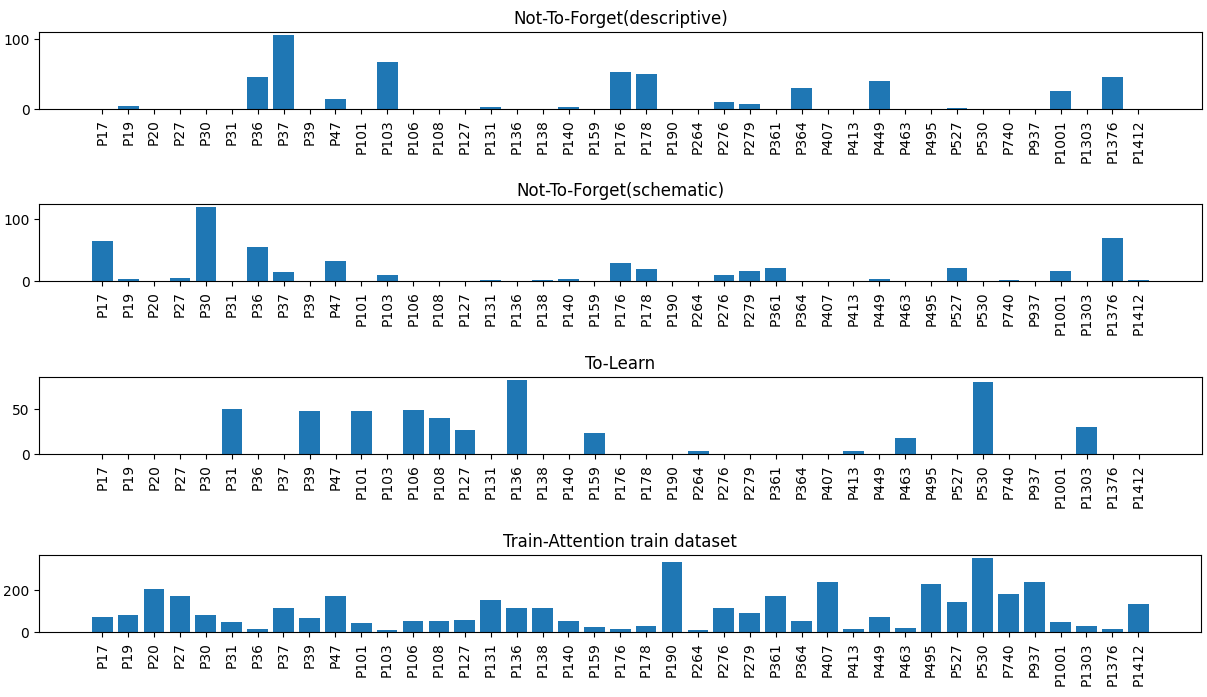}}%
  \caption{Distribution of relation categories of each dataset in \textsc{LAMA-ckl}}
  \label{fig:lama-distribution}
\end{figure}

As each unit of the LAMA dataset contains multiple evidence documents, we sample one document for each unit, based on the following criteria: (1) The document should exceed a length of 70 tokens. (2) The document must include both the subject and object entities. These documents are then truncated to a maximum length of 512 tokens. We assess the accuracy of the knowledge base triple of each unit to filter them. We utilize the TinyLlama-1.1B \cite{zhang2024tinyllama} and Llama2-7B \cite{touvron2023llama} models, integrated with QLoRA \cite{dettmers2024qlora} and K-Adapter \cite{wang2020k}, total 4 variations. Units with unanimous accuracy among the four models are only selected for \textsc{to-learn} or \textsc{not-to-forget} set. Because \textsc{not-to-forget} accuracy differs among descriptive and schematic form queries, \textsc{not-to-forget} set for descriptive and schematic tasks are separately sampled, allowing overlap. The distribution of relation categories among \textsc{to-learn} and \textsc{not-to-forget} is depicted in the Figure~\ref{fig:lama-distribution}. Also the relation categories of each \textsc{to-learn} and \textsc{not-to-forget} are detailed in Table~\ref{tab:relations}. Table~\ref{tab:lama_stat} presents data statistics for \textsc{not-to-forget} and \textsc{to-learn} sets.

\begin{table}[h!]
\centering
\hfill

\caption{\textsc{LAMA-ckl} data statistics.}

\centering
\begin{tabular}{ lccccl}
\toprule
                          & \thead{\textbf{Size}}      &\thead{ \textbf{Avg.} \textbf{Evidence token}}     & \thead{\textbf{Avg.} \textbf{Label token}} \\ \hline
{\textsc{not-to-forget} (descriptive)}           & 500        & 141.1     & 1.5  \\
{\textsc{not-to-forget} (schematic)}                &  500     & 153.4 & 1.9  \\
{\textsc{to-learn}}       &500            &112.8   &1.0    \\
{Train-Attention Train Data}          & 4166       & 136.0  & 1.1 \\
\bottomrule
\end{tabular}
\label{tab:lama_stat}
\end{table}

\begin{table}[h!]
\caption{Relations of each \textsc{not-to-forget} and \textsc{to-learn} set.}
\label{tab:relations}
\begin{subtable}{\linewidth}
\centering
\caption{Relations of \textsc{not-to-forget}\tablefootnote{Directly referred from `Relations of InvariantLAMA of \cite{jang2021towards}}}
\begin{tabular*}{1\columnwidth}{lcccl}
    \toprule
    \textbf{Relation Code} & \textbf{Template ({[}X{]}, {[}Y{]})} & \textbf{Relation} \\ \midrule
    P19 & {[}X{]} was born in {[}Y{]} . & place of birth \\
    P20 & {[}X{]} died in {[}Y{]} . & place of death \\
    P279 & {[}X{]} is a subclass of {[}Y{]}. & subclass of \\
    P37 & The official language of {[}X{]} is {[}Y{]}. & official language \\
    P449 & {[}X{]} was originally aired on {[}Y{]} . & original network \\
    P47 & {[}X{]} shares border with {[}Y{]} . & shares border with \\
    P138 & {[}X{]} is named after {[}Y{]} . & named after \\
    P364 & The original language of {[}X{]} is {[}Y{]} . & original language of film or TV show \\
    P527 & {[}X{]} consists of {[}Y{]} . & has part \\
    P176 & {[}X{]} is produced by {[}Y{]} . & manufacturer \\
    P27 & {[}X{]} is {[}Y{]} citizen . & country of citizenship \\
    P407 & {[}X{]} was written in {[}Y{]} . & language of work or name \\
    P30 & {[}X{]} is located in {[}Y{]} . & continent \\
    P178 & {[}X{]} is developed by {[}Y{]}. & developer \\
    P1376 & {[}X{]} is the capital of {[}Y{]}, & capital of \\
    P131 & {[}X{]} is located in {[}Y{]} . & located in the administrative territorial entity \\
    P1412 & {[}X{]} used to communicate in {[}Y{]}. & languages spoken, written or signed \\
    P17 & {[}X{]} is located in {[}Y{]} . & country \\
    P276 & {[}X{]} is located in {[}Y{]} . & location \\
    P937 & {[}X{]} used to work in {[}Y{]}. & work location \\
    P140 & {[}X{]} is affiliated with the {[}Y{]} religion . &religion \\
    P103 & The native language of {[}X{]} is {[}Y{]} . & native language\\
    P190 & {[}X{]} and {[}Y{]} are twin cities . & twinned administrative body \\
    P1001 & {[}X{]} is a legal term in {[}Y{]} . & applies to jurisdiction \\
    P495 & {[}X{]} was created in {[}Y{]} . & country of origin \\
    P36 & The capital of {[}X{]} is {[}Y{]} . & capital \\
    P740 & {[}X{]} was founded in {[}Y{]}. & location of formation \\
    P361 & {[}X{]} is part of {[}Y{]} . & part of \\
    \bottomrule
    \end{tabular*}
\end{subtable}

\begin{subtable}{\linewidth}

\centering
\caption{Relations of \textsc{to-learn}}
\begin{tabular*}{1\columnwidth}{lcccl}
    \toprule
    \textbf{Relation Code} & \textbf{Template ({[}X{]}, {[}Y{]})} & \textbf{Relation} \\ \midrule
    P31 & [X] is a [Y] . & instance of \\
    P39 & [X] has the position of [Y] . &position held \\
    P101 & [X] works in the field of [Y]. & field of work \\
    P106 & [X] is a [Y] by profession. & occupation \\
    P108 & [X] works for [Y] . & employer \\
    P127 & [X] is owned by [Y] . & owned by \\
    P136 & [X] plays [Y] music . &genre \\
    P159 & The headquarter of [X] is in [Y] . & headquarters location \\
    P264 & [X] is represented by music label [Y]. & record label \\
    P413 & [X] plays in [Y] position . & position played on team / speciality \\
    P463 & [X] is a member of [Y] . & member of\\
    P530 & [X] maintains diplomatic relations with [Y] . &diplomatic relation \\
    P1303 & [X] plays [Y] . &instrument \\
  
    \bottomrule
    \end{tabular*}
\end{subtable}
\end{table}

\newpage
\subsubsection{Evaluation setup detail}
\label{app:lama-eval-detail}
\paragraph{Hardware and hyper-parameters}
During training \textsc{to-learn} documents, 8 RTX 3090 GPU (24GB) are used, with a global batch size of 64. A total of 30 epochs took 25 minutes of GPU time. Learning rate 1e-4, AdamW optimizer, and max length of 512 tokens are applied.

\paragraph{Accuracy measurement} Accuracy is utilized as a metric (Top Acc, NF Acc) to assess the efficacy of models in the task of label prediction. This metric quantifies the proportion of label tokens correctly identified by the model out of the total label tokens presented.

\subsection{Train-Attention training detail}
\label{app:train-attention}
\paragraph{Dataset}
We select LAMA units from both time-variant and time-invariant sets, specifically those with an accuracy below 0.5, while ensuring no overlap with the \textsc{not-to-forget} and \textsc{to-learn} sets. Furthermore, the initial data exhibit a disproportionately large number of units categorized under relation P530 (`diplomatic relations of the country'). To address this imbalance, we adjust the frequency of P530 units to match that of the second most prevalent type. Consequently, the final version of the Train-Attention training dataset comprises a total of 4166 units, detailed in Table~\ref{tab:lama_stat}.

We propose the \textbf{schematic form} to arrange the knowledge base triple into a human-readable sentence. It fits the one-directional feature of causal LM, as opposed to the descriptive form where important information sometimes appears behind the label tokens. The template and example are on Appendix~\ref{app:desc_schem}. We employ schematic form in training Train-Attention, while employing descriptive form during evaluation.

\paragraph{Training}
We utilized small (TinyLlama-1B) and large (Llama2-7B) models from the LLaMA family, integrated with QLoRA \citep{dettmers2024qlora}. We found that using a large rather than small model for the base model ($\theta$) results in faster convergence. On the other hand, employing a large or a small model for the Train-Attention model ($\phi$) made little difference in the aspect of the convergence step and validation score. So we adopt a large model for $\theta$ and a small model for $\phi$ while training Train-Attention. In the test phase, We find that $\theta$ and $\phi$ are still compatible even though they don't share the same background model, as long as they use the same tokenizer. Therefore we utilize small-size Train-Attention for both small and large baseline experiments.  

We initialize the parameters of the Train-Attention head, which is a decoder layer as depicted in Figure~\ref{fig:architecture_inference}a, using a normal distribution with a mean of 0 and a standard deviation of 0.0001. Consequently, the weights generated by the initialized Train-Attention are equivalent to the uniform weight where all \(w_i = 1\). This approach is adopted to observe the convergence of the $\mathcal{L}_{\theta'}(\mathcal{T_D})$ more clearly, ensuring that the loss value declines from the initial state if training proceeds normally.

A single A100 (82GB) GPU is used, and the effect of batch size 16 is achieved through gradient accumulation. A checkpoint of 400 global steps is used (takes about 6 GPU hours). Of the 4166 train data, 100 are used for validation. The learning rate of 2e-4 and AdamW optimizer is employed for both the base model and Train-Attention.

\subsection{Baselines detail}
\label{app:lama-baselines}
In this section, we describe the detailed configuration of QLoRA and K-Adapter. The other baselines follow the previous CKL work of \citep{jang2022temporalwiki}.
\paragraph{QLoRA} Hyper-parameters of QLoRA follow one of the optimal of the original work \citep{dettmers2024qlora}. We employ LoRA $r=64, \alpha=16$, NF4 with BF16 computation datatype. A total of 160M parameters are expanded for the large (Llama2-7B) baselines.

\paragraph{K-Adapter} Hyper-parameters of K-Adapter follow the previous CKL work of \citep{jang2022temporalwiki}. A total of 303M parameters are expanded. The trainable parameters are double the QLoRA. Efforts are made to reduce the learnable parameters of K-Adapter, but we observe that the current settings are the minimum level necessary for K-Adapter to function effectively. In experiments, K-Adapter demonstrates greater stability than QLoRA, which could be partially attributed to its larger parameter size. We also employ the same quantization configuration of QLoRA for K-Adapter, NF4 with BF16 computation datatype.

\subsection{Experimental results additional detail}
\subsubsection{\textsc{LAMA-ckl} Experiment on small (1B) baselines}
\label{app:lama-1b}
\begin{figure}[h!]
 \begin{subfigure}{0.5\textwidth} 
    \centering
    \includegraphics[width=\linewidth]{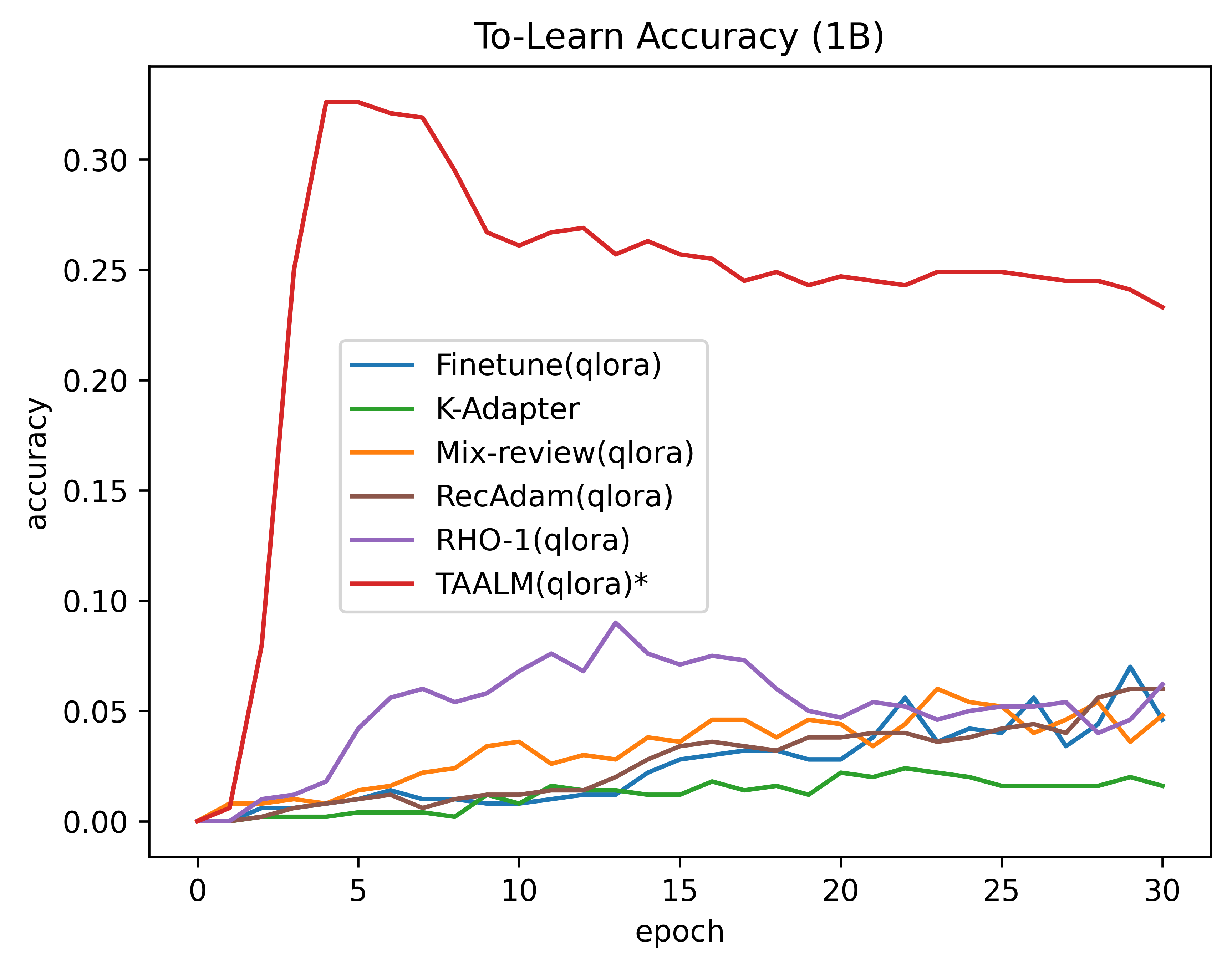}
    \end{subfigure}%
 \begin{subfigure}{0.5\textwidth} 
    \centering
    \includegraphics[width=\linewidth]{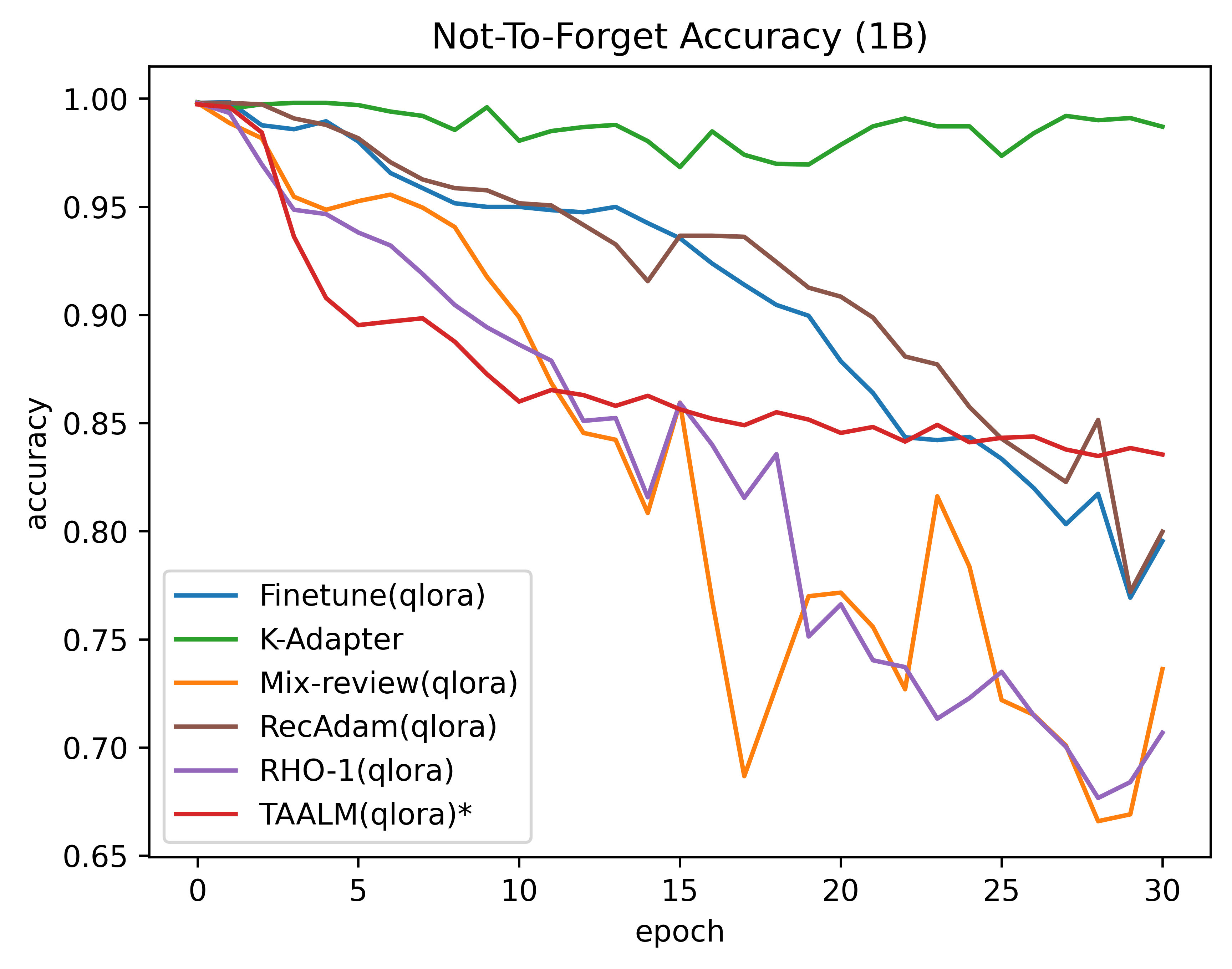}
    \end{subfigure}%
  
  \caption{\textsc{LAMA-ckl} performance of small (TinyLlama-1B) baselines.}
  \label{fig:lama-1b}
\end{figure}

\autoref{fig:lama-1b} and \autoref{tab:lama_1b} describe performances of small (TinyLlama-1B) baselines on the \textsc{LAMA-ckl}. The performance of TAALM based on the small (TinyLlama-1B) architecture is significantly better even when compared to the large (Llama2-7B) baselines.

\begin{table}[h!]
\hfill
\caption{\textsc{LAMA-ckl} performance of small (TinyLlama-1B) baselines.}
\label{tab:lama_1b}

\centering
\begin{tabular}{
 lllll
}
\toprule
                          & \thead{Top Acc}      & Epoch     & \thead{NF Acc} &\thead{Total Knowledge} \\ \hline
Finetune(QLoRA)           & 0.0700        & 29     & 0.7693   & 0.8393 \\
K-Adapter                 &  0.0240     & 22 & \textbf{0.9908} & \underline{1.0148} \\
Mix-review(QLoRA)        &0.0600            &23   &0.8161 & 0.8761   \\
RecAdam(QLoRA)           &0.0600            &29   & 0.7719 &0.8319  \\
RHO-1(QLoRA)             & \underline{0.0900}      & \underline{13 }  & 0.8523  & 0.9423\\  \hdashline \addlinespace
TAALM(QLoRA)           & \textbf{0.3260}      & \textbf{4}  & \underline{ 0.9078} & \textbf{1.2338}  \\
\bottomrule
\end{tabular}
\end{table}

\subsubsection{Combination of ours (TAALM) and other baselines}
\label{app:combination}
\begin{figure}[h!]
  \centerline{

 \begin{subfigure}{0.34\textwidth} 
    \centering
    \includegraphics[width=\linewidth]{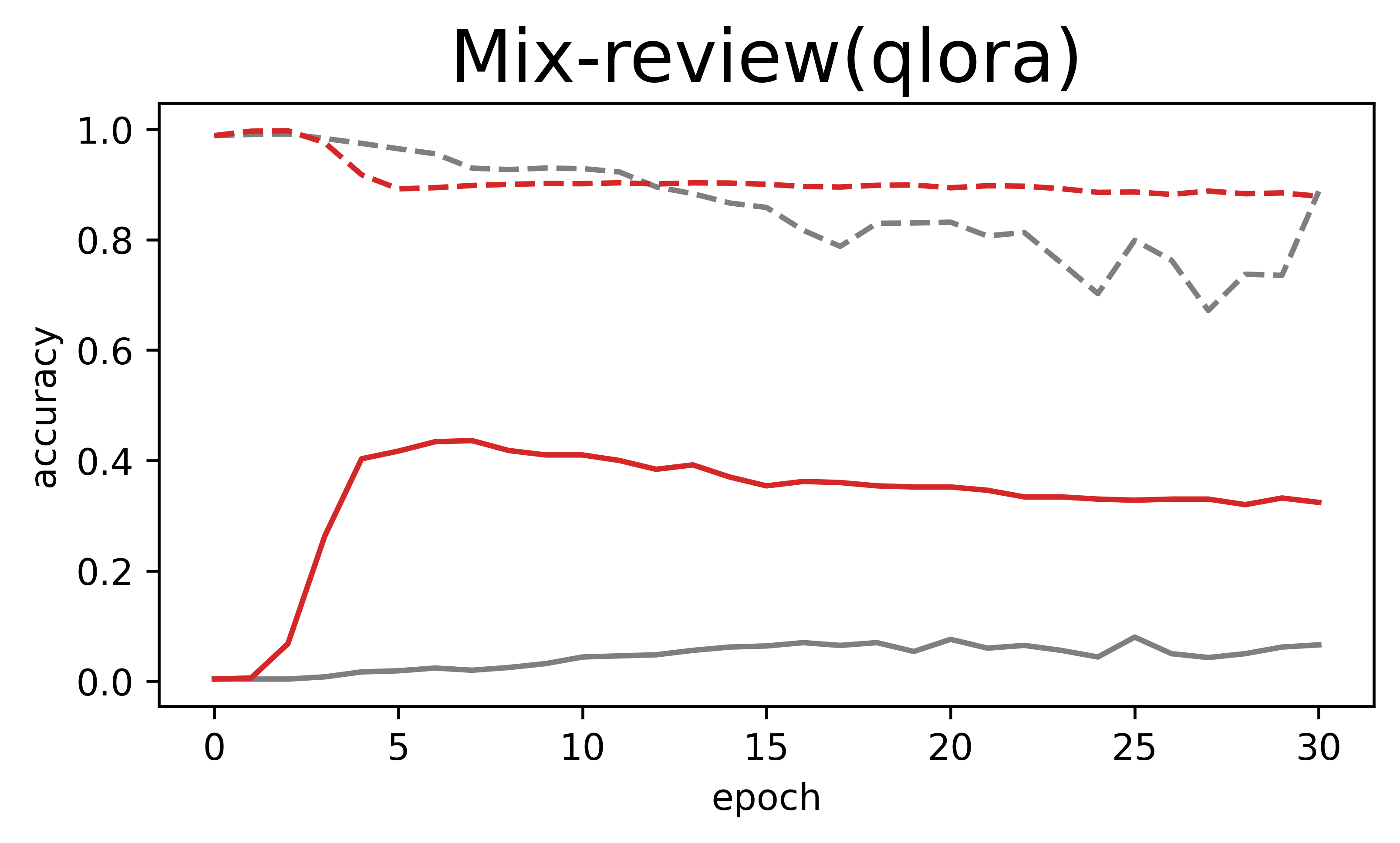}
    \end{subfigure}%
 \begin{subfigure}{0.34\textwidth} 
    \centering
    \includegraphics[width=\linewidth]{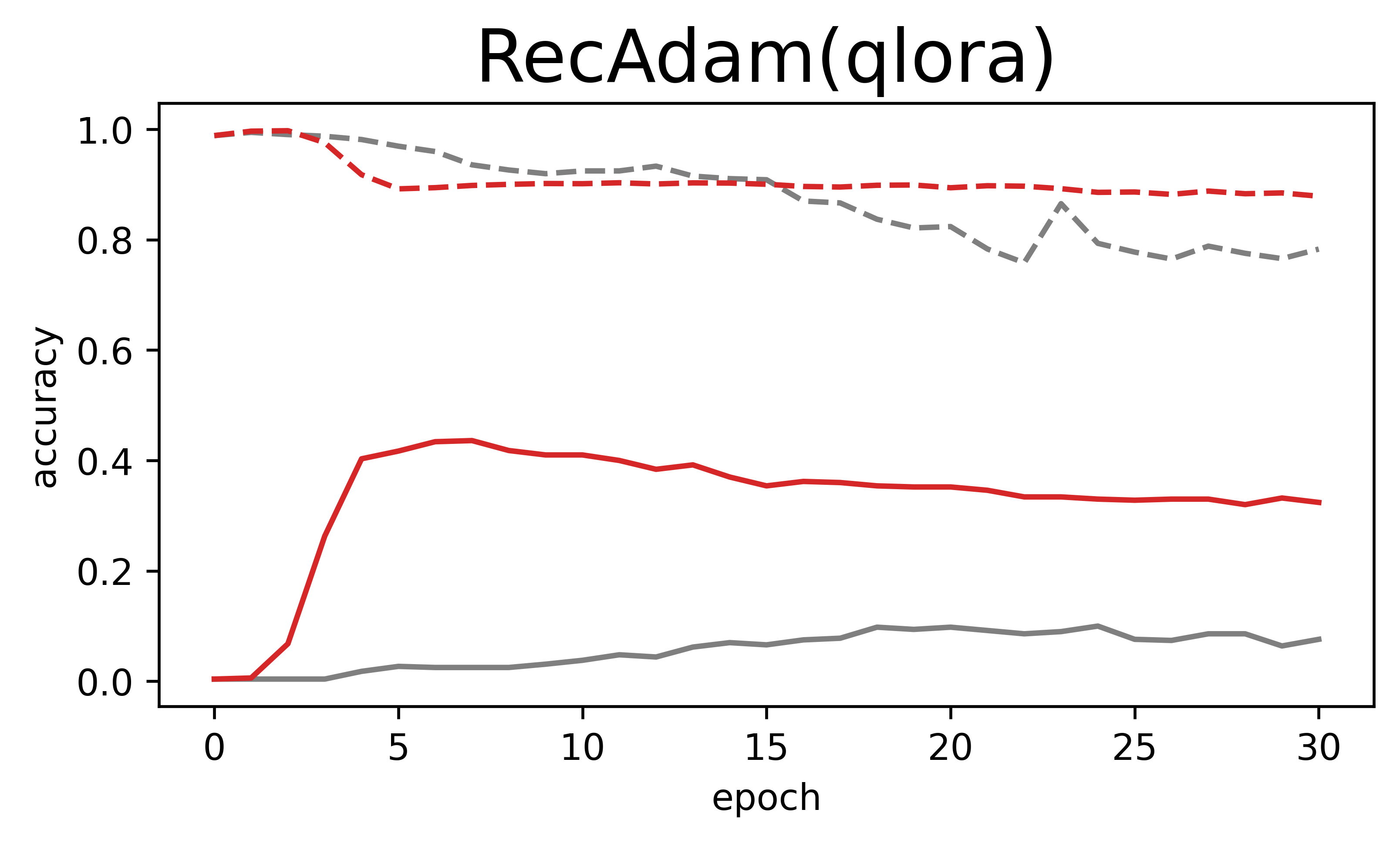}
    \end{subfigure}%
 \begin{subfigure}{0.34\textwidth} 
    \centering
    \includegraphics[width=\linewidth]{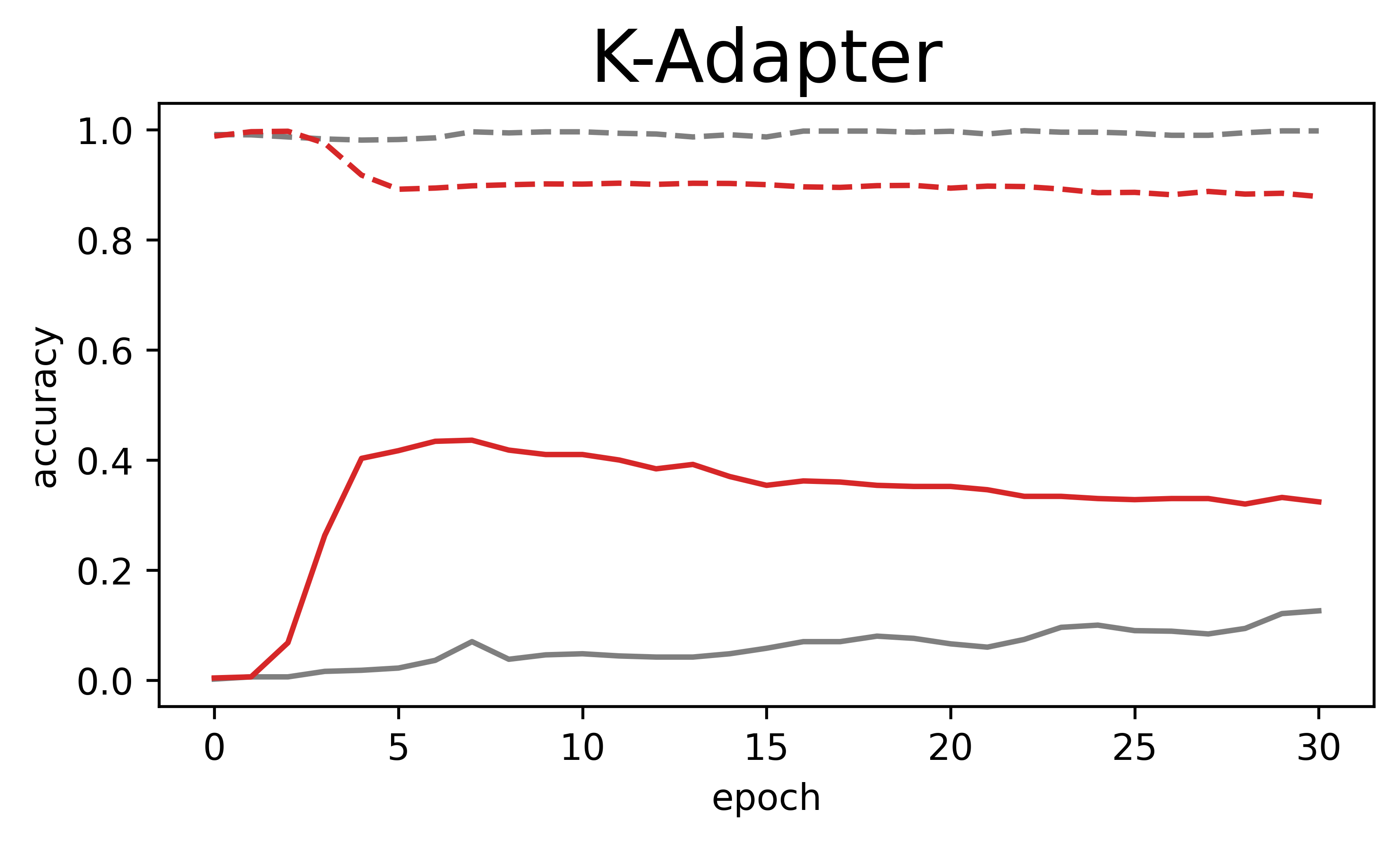}
    \end{subfigure}%
    }

  \caption{Comparison of each baseline alone and combined with our method. Each title on the plot represents the baseline method. The {\color{gray}\textbf{gray line}} represents the baseline alone, and the {\color{red}\textbf{red line}} represents the combination with TAALM. Solid line for \textsc{to-learn}, dashed line for \textsc{not-to-forget}. All are based on Llama2-7B, and tested on \textsc{LAMA-ckl}. }
  
  \label{fig:combine}
\end{figure}

Figure~\ref{fig:combine} presents that the performances of all baselines are significantly improved when combined with the Train-Attention, indicating synergistic compatibility of our method.

\begin{figure}[h!]
  
 \begin{subfigure}{0.5\textwidth} 
    \centering
    \includegraphics[width=\linewidth]{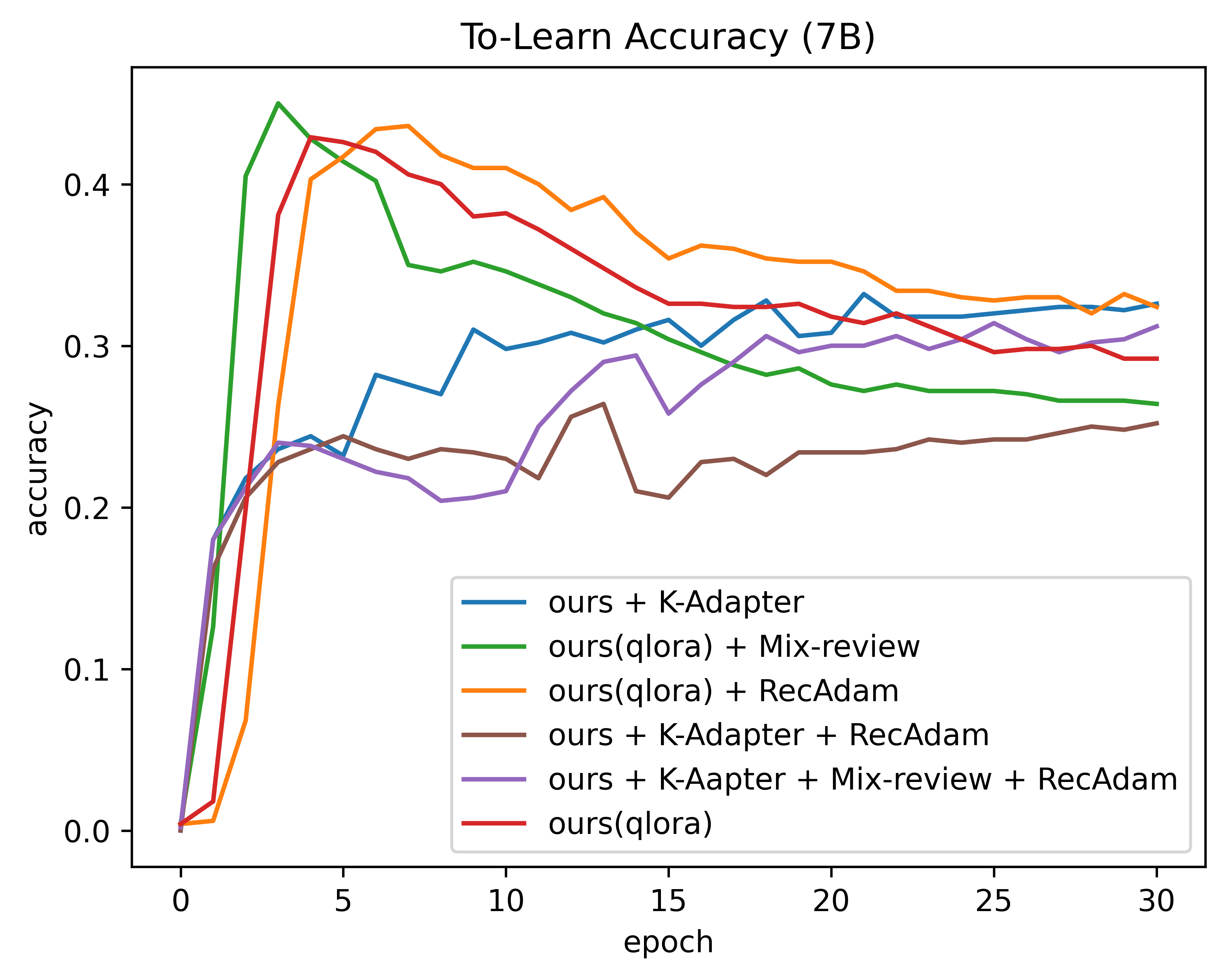}
    \end{subfigure}%
 \begin{subfigure}{0.5\textwidth} 
    \centering
    \includegraphics[width=\linewidth]{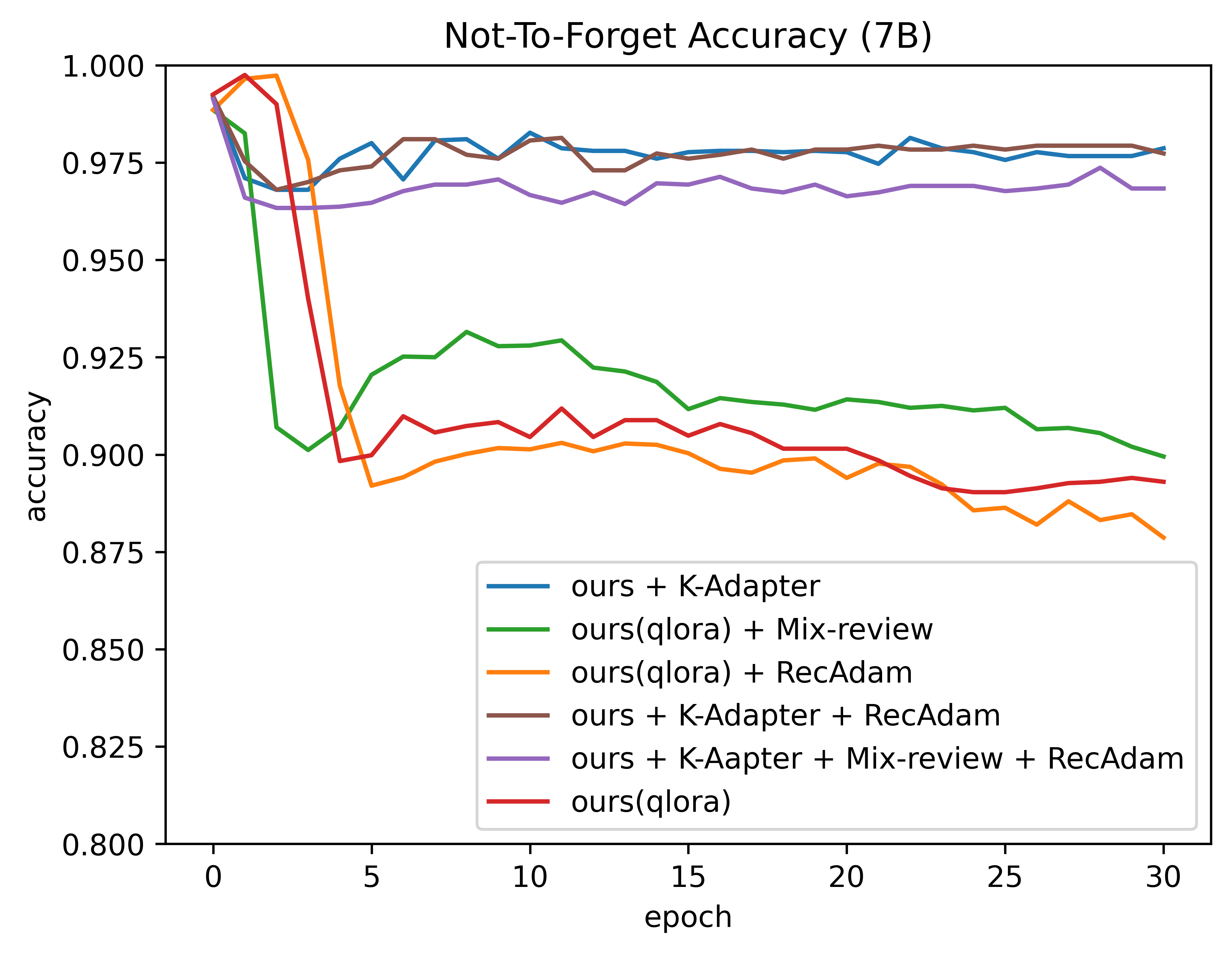}
    \end{subfigure}%
  \caption{Combinations of other methods with ours (TAALM) on Llama2-7B base model, tested on \textsc{LAMA-ckl}.}
  \label{fig:lama-ablation}
\end{figure}

Referring to Table~\ref{tab:lama-ablation} and Figure~\ref{fig:lama-ablation}, the combination of ours and Mix-review shows the highest Top Acc and Total Knowledge among all variations. Unlike Mix-review alone, the Mix-review combined with ours applies Train-Attention not only to the train data but also to the review data. As a result, it is observed that the training becomes more stable than the original, which tends to be unstable due to the doubled amount of data (train + review). This indicates that the combination of Train-Attention with the rehearsal approach is synergistic. Variations of ``ours + K-Adapter'' show relatively low learning capability (Top Acc, Total Knowledge) while featuring still robust stability. Therefore, a combination of ours and K-Adapter can be considered a balanced variation between learning and maintaining.

\begin{figure}[h!]
  \centerline{
 \begin{subfigure}{0.5\textwidth} 
    \centering
    \includegraphics[width=\linewidth]{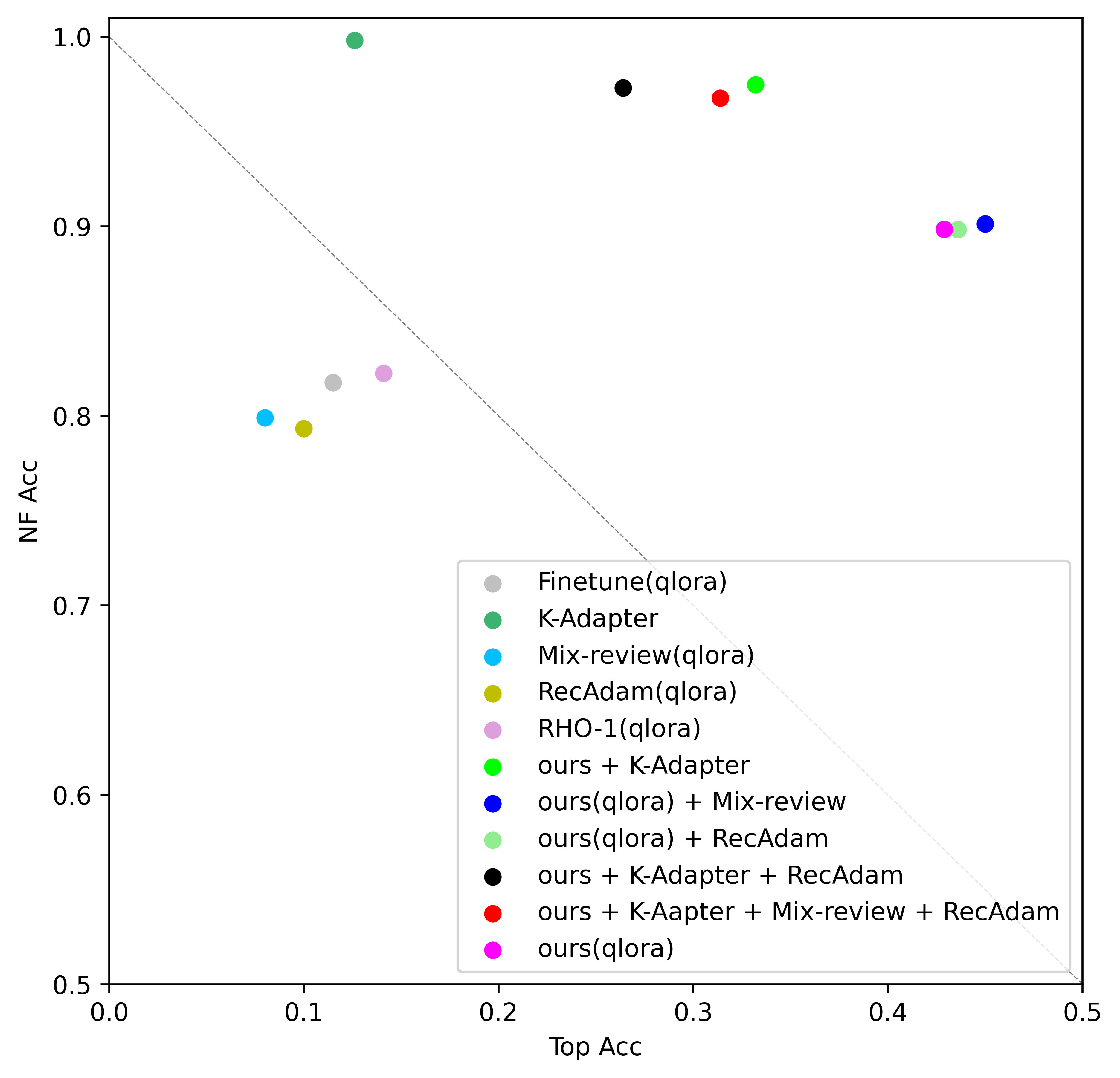}
    \caption{Dots}
    \end{subfigure}%
 \begin{subfigure}{0.5\textwidth} 
    \centering
    \includegraphics[width=\linewidth]{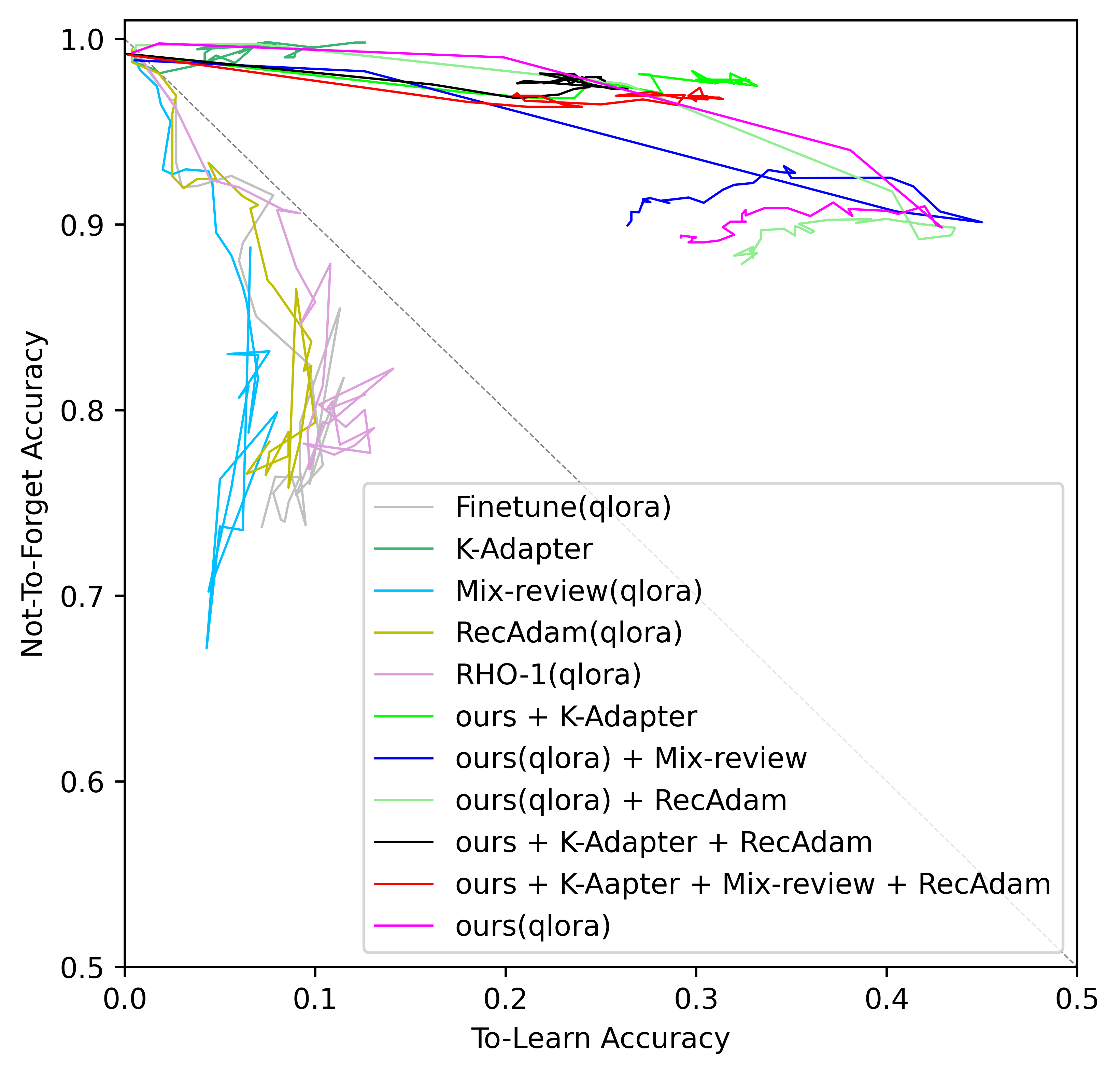}
    \caption{Lines}
    \end{subfigure}%
    }

  \caption{The trade-off between plasticity (\textsc{to-learn}) and stability (\textsc{not-to-forget}) are visualized, for all baselines including combinations with our method. All models are based on large (Llama2-7B) architecture. Gray dashed lines stand for the zero-sum state where the Total Knowledge (sum of the performances of \textsc{to-learn} and \textsc{not-to-forget}) is 1. The right and upper sides of the gray lines indicate the more efficient system where learning causes less forgetting. \textbf{(a)Dots} presents only the checkpoints of the Top Acc, while \textbf{(b)Lines} presents whole checkpoints of 30 epochs as lines.}
  
  \label{fig:lama-variation}
\end{figure}

Figure~\ref{fig:lama-variation} presents the trade-off between plasticity (\textsc{to-learn}) and stability (\textsc{not-to-forget}) of all baselines, including all variations of combination. Variations combined with our method are exclusively positioned in the upper right quadrant, indicating a minimal trade-off between updating and forgetting.

\begin{table}[h!]

\caption{Combination of ours (TAALM) and other baselines. Based on Llama2-7B, tested on \textsc{LAMA-ckl}.}
\hfill

\label{tab:lama-ablation}

\hfill

\centering
\resizebox{\textwidth}{!}{\begin{tabular}{
 lllll
}
\toprule
                          & Top Acc      & Epoch     & NF Acc &Total Knowledge   \\ \hline
ours + K-Adapter           &  0.3320      & 21   & \textbf{0.9747}   & 1.3067 \\
ours(QLoRA) + Mix-review  &  \textbf{0.4500}  & \textbf{3} & 0.9012 & \textbf{1.3512} \\
ours(QLoRA) + RecAdam  & \underline{0.4360}  & 7   &0.8982 & \underline{1.3342}   \\
ours + K-Adapter + RecAdam    &0.2640            &13   & \underline{0.9730} &1.2370  \\
ours + K-Adapter + Mix-review + RecAdam   & 0.3140      & 25  & 0.9677  & 1.2817\\  \hdashline \addlinespace
ours(QLoRA)     & 0.4290     & \underline{4}  &0.8983 & 1.3273  \\
\bottomrule
\end{tabular}}
\end{table}

\newpage

\subsubsection{Rationale for the accuracy drop}
In the main experiment (\autoref{fig:lama_7b}), the accuracy declines after a rapid peak after about the fourth epoch. We hypothesize this is a result of overfitting. While the token weights from TA include beneficial targets, there must also be some that are not. Performance achieves peak until the model completes learning for the true target, but learning may continue for the false targets. This continued learning leads to parameter updates in suboptimal directions, resulting in forgetting.

The comparison between TA and the oracle (\autoref{fig:oracle}) provides evidence for this rationale. As the oracle's performance does not decline with further training, the differences in accuracy trends could be caused by false targets. This phenomenon can also be interpreted as a type of overfitting where the model becomes too fitted to the training data which has different distribution from test data. As the phenomenon of declining after peaking is common due to overfitting, this cycle seems to occur more rapidly for TAALM.

\subsubsection{Reporting error ranges of the main experiment}
\label{app:error_bar}
\begin{figure}[h!]
  \centerline{
 \begin{subfigure}{0.5\textwidth} 
    \centering
    \includegraphics[width=\linewidth]{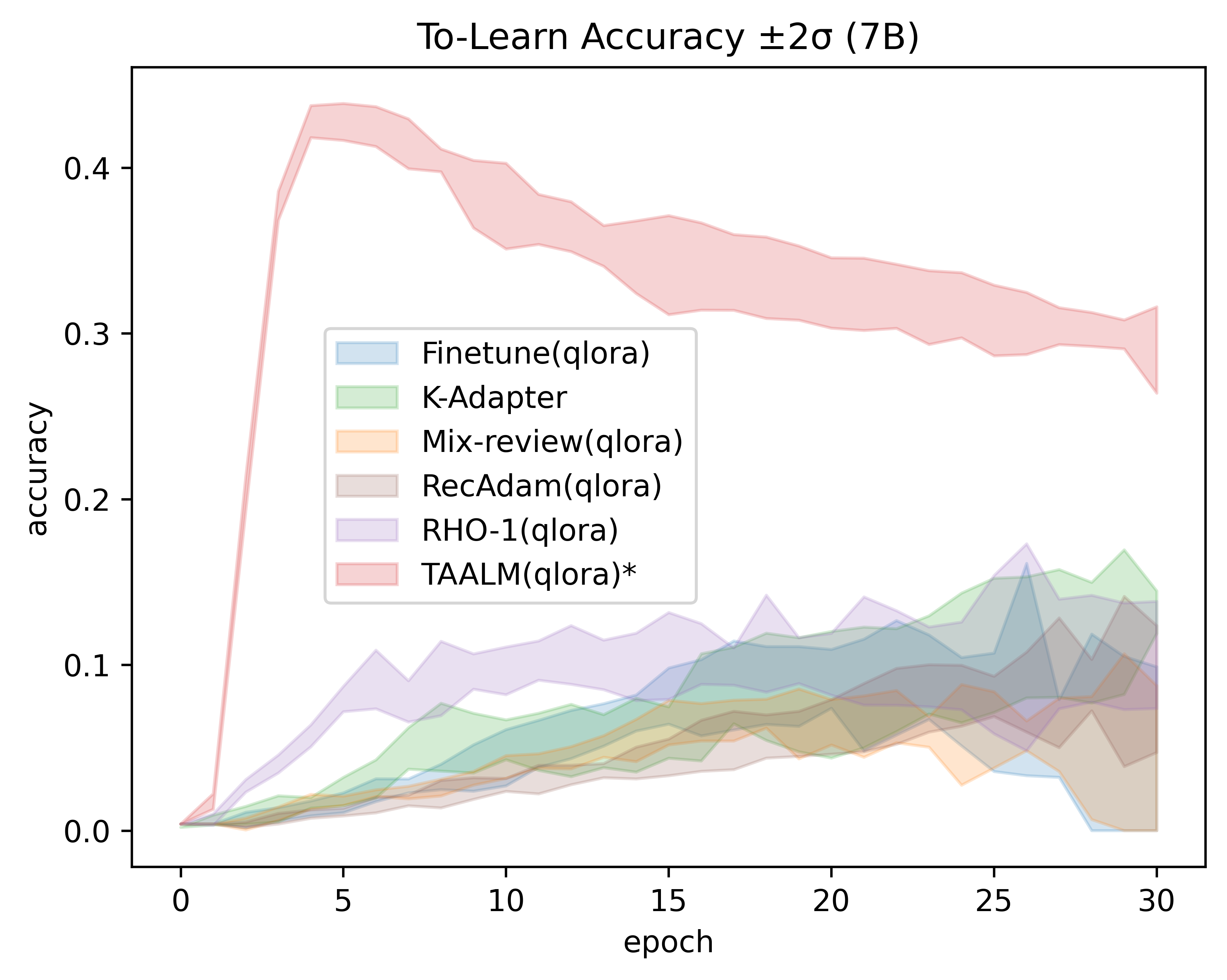}
    \caption{Dots}
    \end{subfigure}%
 \begin{subfigure}{0.5\textwidth} 
    \centering
    \includegraphics[width=\linewidth]{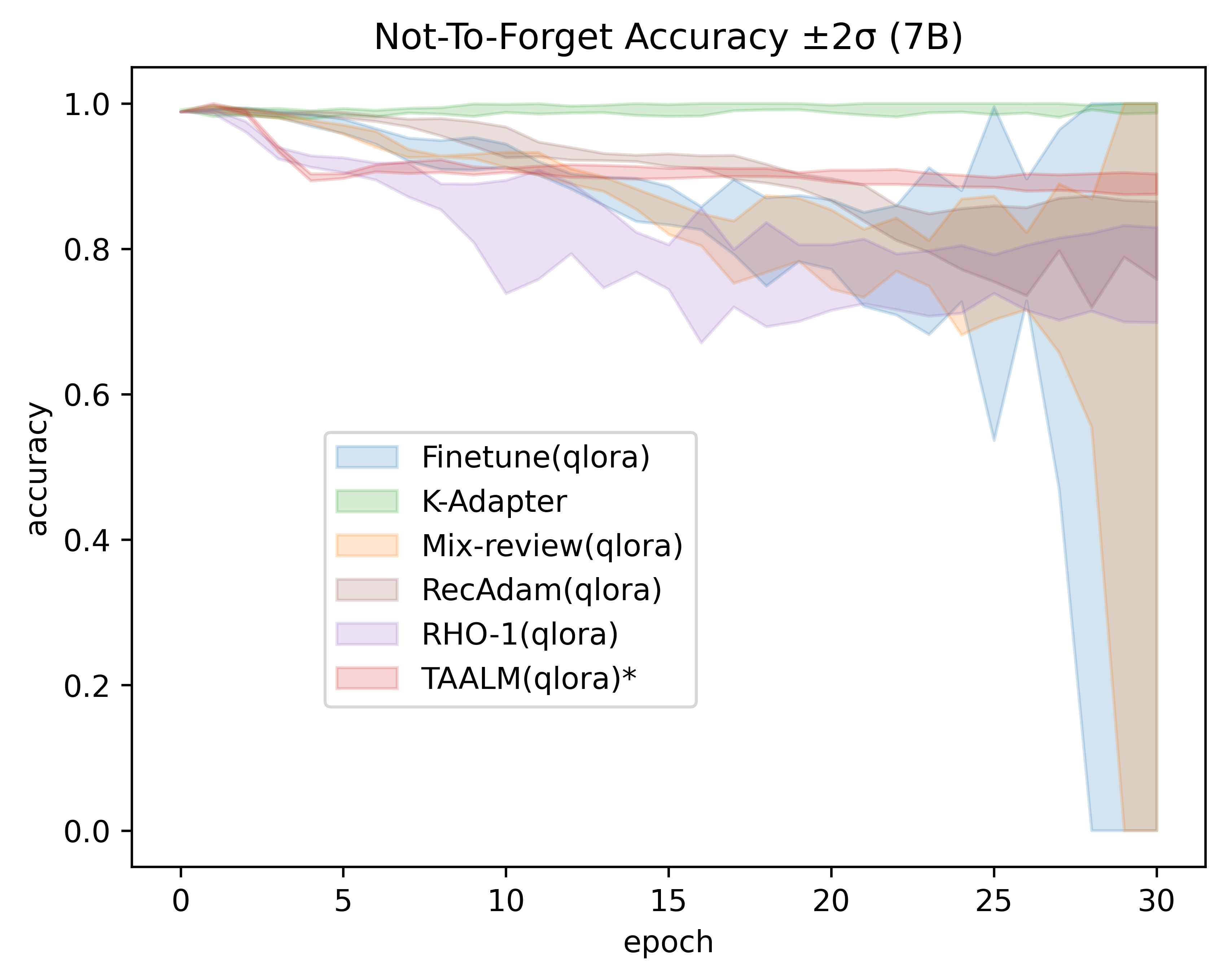}
    \caption{Lines}
    \end{subfigure}%
    }

    \caption{Performance of the large-scale baseline models (Llama2-7B) on LAMA-CKL, depicted with $\pm 2\sigma$ error ranges. Results are calculated from five random trials, employing different random seeds over the train data-loader.}
  \label{fig:lama-sigma}
\end{figure}

We conduct five independent runs using different random seeds on the training data-loader, and display the results with error ranges at $\pm 2\sigma$ in Figure~\ref{fig:lama-sigma}. This demonstrates the consistent performance of our method. All main experiments presented use the data loader with random seed 42.
{\color{white} d}
\newpage
\section{Descriptive Form \& Schematic Form}
\label{app:desc_schem}
\begin{figure}[h!]
  \makebox[\textwidth][c]{\includegraphics[width=1.01\textwidth]{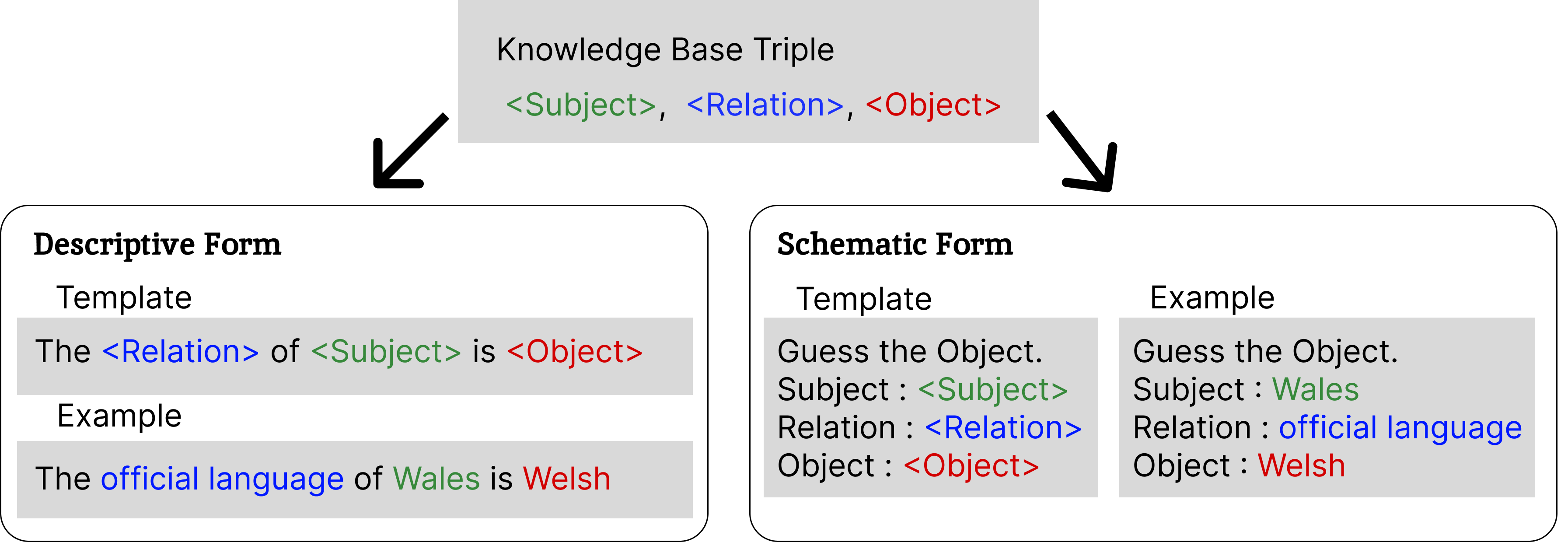}}%
  \caption{Descriptive form and our proposed schematic form template to rearrange knowledge triple into a human readable document.}
\label{fig:desc_schem}
\end{figure}

Each unit of the LAMA dataset comprises a knowledge base triple and a corresponding sentence that encapsulates the information contained within the triple. These sentences are presented in a short \textit{descriptive form}. In contrast, our newly proposed \textbf{schematic form} organizes these triples more systematically. The template and example are described in Figure~\ref{fig:desc_schem}. We propose the schematic form because it better aligns with the uni-directional nature of causal language models (LMs). In the descriptive form, critical information often comes after the label tokens, which can be problematic. For instance, in the descriptive form template "[X] is [Y] citizen", the causal LM lacks the crucial cue `citizen' when predicting [Y]. This not only makes the assessment less accurate but also introduces noise into the train-attention learning process.

\section{Detail of \textsc{TemporalWiki} Experiment}
\label{app:temporalwiki}
\subsection{Corpus setup}
The original corpus consists of two datasets to learn for 4 periods (0809, 0910, 1011, 1112); (1) Wikipedia: the whole snapshot of Wikipedia documents of each period. (2) \textsc{Diffset}s: the segments of documents that only contain changed information compared to the last period snapshot. The corpus also includes test datasets (TWiki-Probes) corresponding to each period. Just like the \textsc{LAMA-ckl}, the Wikipedia snapshots and \textsc{Diffset}s consist of evidence documents, while TWiki-Probes consist of knowledge base triples. Because we used the data from the very first period to train Train-Attention, only the data from the next three periods (0910, 1011, 1112) are used for the evaluation.

The original study shows that learning only the \textsc{Diffset} results in better performance than learning the entire snapshot. So we excluded the baselines that learn full snapshots and set learning of \textsc{Diffset} as the default. We also employed heuristic filtering for the \textsc{Diffset}. Empty texts containing only `nan' in the content, or texts with more than 70\% of non-letters, which are assumed to contain little meaningful information, are filtered out. Therefore the total number per \textsc{Diffset} data decreases from an average of 837K to 707K documents per each dataset.

Each Twiki-Probes consists of two parts: (1) \textsc{Changed}: assumed to consist only of knowledge base triples that contain changed information compared to the last period. (2) \textbf{Unchanged}: contain only retained information from the last period, which is the complementary set of the \textsc{Changed}. \textsc{Changed} set is supposed to measure plasticity, while \textsc{Unchanged} set is supposed to measure stability.

\subsection{Evaluation and baseline setup}
We follow the evaluation settings of the original work \citet{jang2022temporalwiki}: a model continually learns train datasets of each period while evaluating performance on TWiki-Probes after training of each period. The evaluation metric is the mean perplexity of object label words. One dataset is updated for only 1 epoch.

The original study simply arranges the triples in order (e.g., ``Wales official language Welsh'') and sets the task as predicting the object label token. As this form of sentence might seriously violate the grammar, it has been reported that perplexity tends to pick extraordinarily high in the original work. To handle this, we adopt the schematic task format used in \textsc{LAMA-ckl} (\autoref{lama-ckl-eval-setup}).

Baselines and hyper-parameters settings follow that of \textsc{LAMA-ckl} (\autoref{lama-ckl-eval-setup}), which follow the \citet{jang2022temporalwiki}. 
 
\subsection{Train-Attention training setup}

The \textsc{Diffset} and Twiki-Prob of the first period (0809) are used for training Train-Attention.
To train Train-Attention (Algorithm~\ref{alg:cap}), data ($\mathcal{D}$) and the corresponding task ($\mathcal{T_D}$) must be paired, while the documents in \textsc{Diffset} and knowledge base triples in Twiki-Probes are not paired. To tackle this challenge, we search the \textsc{Diffset} documents which contain both the object and subject of the Twiki-Probes unit and manually join them. The total number of pairs is 5235. Validation setting and hyper-parameters follow \autoref{lama-ta-train-setup}.

\subsection{Analysis on the results}
\label{app:twiki-anal}
\begin{figure}[H]
  
 \begin{subfigure}{0.5\textwidth} 
    \centering
    \includegraphics[width=\linewidth]{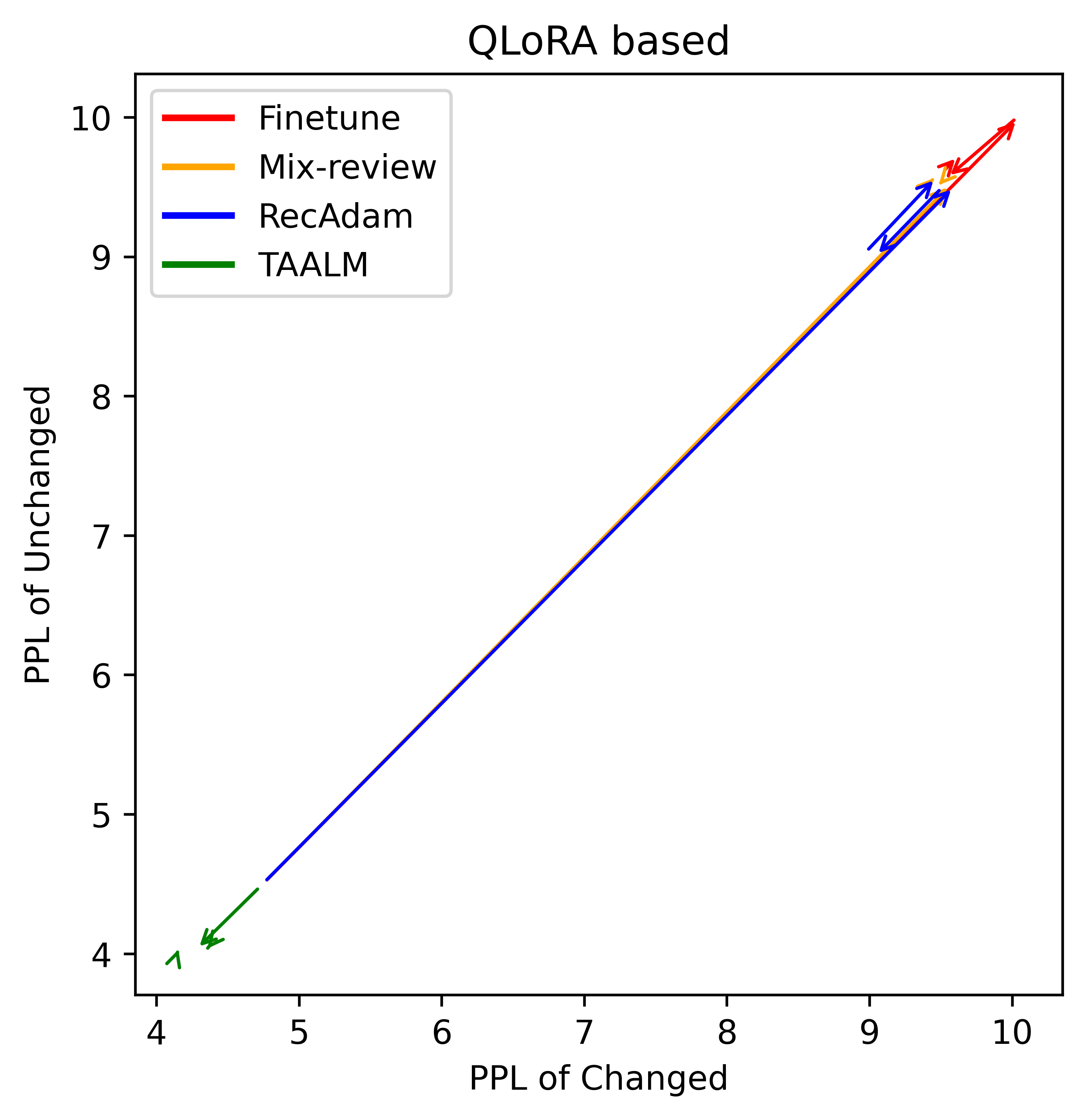}
    \end{subfigure}%
 \begin{subfigure}{0.5\textwidth} 
    \centering
    \includegraphics[width=\linewidth]{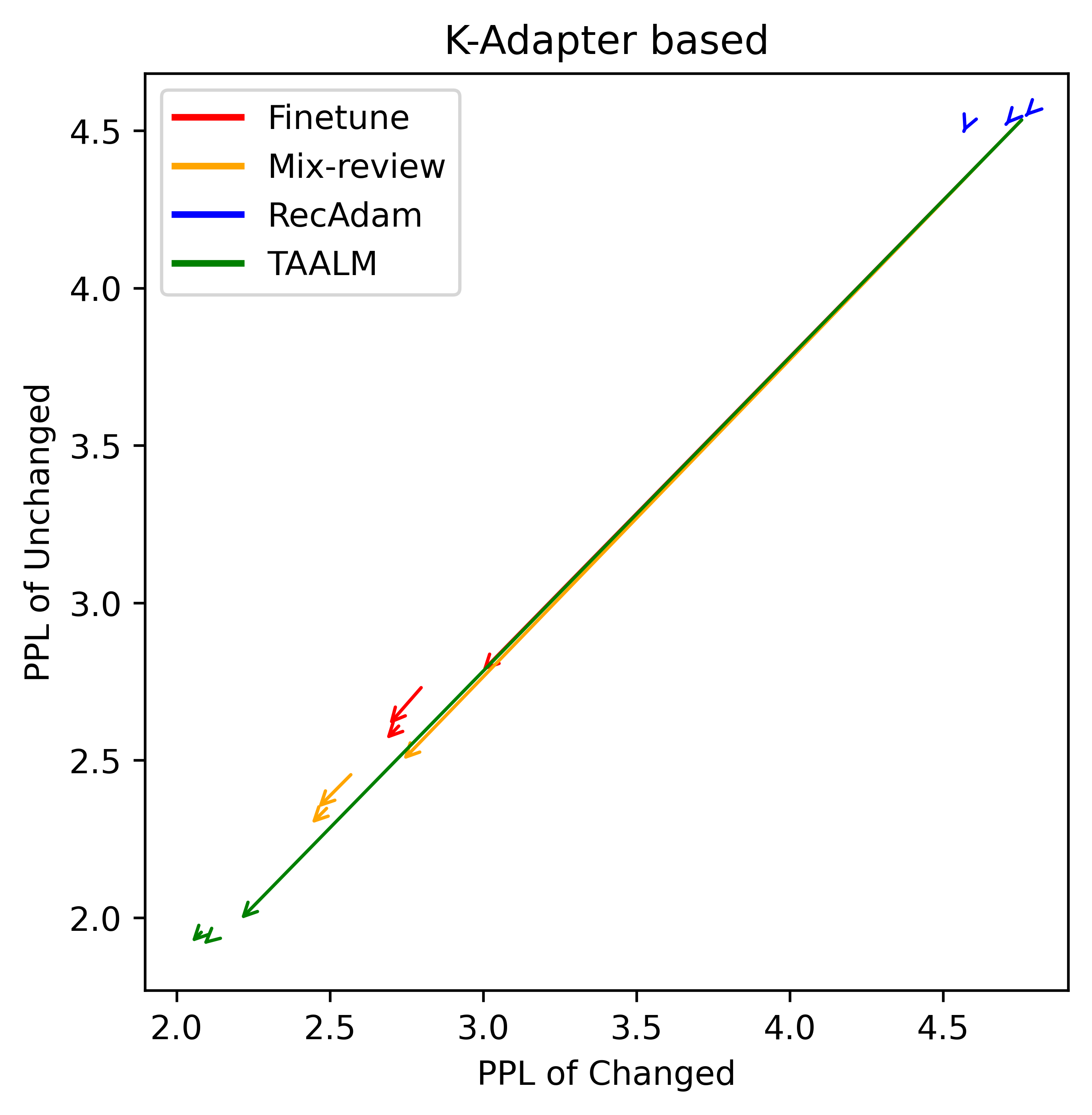}
    \end{subfigure}%
  \caption{The experimental result of small(TinyLlama-1B) baselines on \textsc{TemporalWiki} benchmark. The x-axis and y-axis correspond to the perplexity of Changed and Unchanged sets, respectively. Each color represents baselines, and the arrows represent the experimental result from each period. The start point of the arrow indicates the performance before training, while the endpoint indicates the performance after training.}
  \label{fig:twiki-vector}
\end{figure}

In this section, we present the detailed experimental result on \textsc{TemporalWiki}, specifically involving the discussion in \autoref{sec:why_lama}, on the reason we propose the new benchmark \textsc{LAMA-ckl}. 
\autoref{fig:twiki-vector} indicates that performances in both Changed and Unchanged mostly move in the same direction when the baselines are trained on the \textsc{Diffset} of each period, presenting a rare trade-off between plasticity (Changed) and stability (Unchanged). This is contrasting to the observation on our new benchmark, \textsc{LAMA-ckl} (\autoref{fig:lama-variation}), which exhibits a clear trade-off between plasticity (\textsc{to-learn}) and stability (\textsc{not-to-forget}). This suggests that the \textsc{LAMA-ckl} is a better benchmark for observing the trade-off between plasticity and stability.

\autoref{tab:twiki-overlap} presents the portion of task sentences in each Changed and Unchanged set, which have evidence documents in each period of the training dataset (\textsc{Diffset}). An average of 8.0\% of every Unchanged set have evidence documents in the \textsc{Diffset} of the corresponding period, indicating that training on \textsc{Diffset} partially results in an improvement in the performance on the \textsc{Unchanged} set. The Unchanged set from the last period (TWiki-Probes-1112) contains evidence documents in the \textsc{Diffset}s from all three preceding periods, suggesting that its supportive effect is cumulative. This is a complication factor for observing stability.

\begin{table*}[h!]
\caption{The portion of units in each period of TWiki-Probes, which have corresponding evidence document in the \textsc{Difffset}. \textbf{Un} and \textbf{C} indicate Unchanged and Changed set, respectively.}
 \resizebox{\textwidth}{!}{\begin{tabular}{ccccccccccc}
    \toprule
    \multicolumn{1}{l}{} & \multicolumn{2}{c}{TWiki-Probes-0910} & \multicolumn{2}{c}{TWiki-Probes-1011} &
    \multicolumn{2}{c}{TWiki-Probes-1112}
    \\ \cmidrule(lr){2-3} \cmidrule(lr){4-5} \cmidrule(lr){6-7}  &  \textbf{Un} & \textbf{C}  & \textbf{Un} & \textbf{C}  & \textbf{Un} & \textbf{C}  \\
    \midrule
\textsc{Diffset-0910}           & 7.8\%      & 83.6\%  & 7.6\%  & 39.5\%     & 7.7\%  & 25.3\%      \\
\textsc{Diffset-1011}         & -      & -         & 7.4\%    & 81.8\%      & 7.9\% & 26.2\%      \\
\textsc{Diffset-1112}         & -      &-         & -      & -        & 8.7\% & 83.3\%    \\
\bottomrule

\end{tabular} }
\label{tab:twiki-overlap}
\end{table*}

{\color{white}d}

{\color{white}d}

{\color{white}d}

{\color{white}d}

{\color{white}d}

\section{ Reduction of resources through TA on BERT}
\label{app:bert_taalm}
Due to the substantial GPU resources required to train the TA, it is necessary to find ways to reduce resource consumption. A promising approach is utilizing Bidirectional Transformer (BERT) as a body for TA, which has high inferential capabilities even at a very small size (108M) compared to the previous body (Tinyllama 1.1B), due to its bidirectional property. Since BERT has a different tokenizer from our generation model, the Llama family, we integrate BERT with the Llama2 tokenizer and pre-train it for one epoch on 17GB Wikipedia documents (9 days using 8 of 24GB GPUs). Then, we finetune this BERT as TA, paired with the generation model of 1B (Tinyllama). This very lightweight TAALM is sufficiently trained on only a single 24GB GPU, significantly reducing resource use compared to the previous version (single 82GB GPU), thus making it affordable for the general environment. On the inference, the TA on BERT demonstrates compatibility with both the 1B and 7B generation models. Although its performance is below that of the TA on Llama, it still exhibits the highest performance among the other baselines.

\begin{table*}[h!]
\caption{\textsc{LAMA-ckl} performance of small (TinyLlama-1B) baselines.}

\label{tab:bert_taalm}
\begin{subtable}{\textwidth}
\subcaption{Baselines with large generation model (Llama2 7B)}
\centering
\begin{tabular}{
 llllll
}
\toprule
                     & \thead{Parameter size of TA}     & \thead{Top Acc}      & Epoch     & \thead{NF Acc} &\thead{Total Knowledge} \\ \hline
Finetune             & NA    & 0.1150        & 16     & 0.8174   & 0.9324 \\
TA (Llama)            & 1.1B & \textbf{0.4290}      & \textbf{4 }  & \underline{0.8983}  & \textbf{1.3273} \\  \hdashline \addlinespace
\textbf{TA (BERT) }           & 108M  & \underline{0.3210}      & \underline{6}  & \textbf{ 0.9388} & \underline{1.2598}  \\
\bottomrule
\end{tabular}
\end{subtable}

\begin{subtable}{\textwidth}
\vspace{2mm}
\subcaption{Baselines with small generation model (Tinyllama 1B)}
\centering
\begin{tabular}{
 llllll
}
\toprule
                     & \thead{Parameter size of TA}     & \thead{Top Acc}      & Epoch     & \thead{NF Acc} &\thead{Total Knowledge} \\ \hline
Finetune             & NA    & 0.0700        & 29     & 0.7693   & 0.8393 \\
TA (Llama)            & 1.1B & \underline{0.3260}      & \textbf{4 }  & \underline{0.9078}  & \textbf{1.2338}\\  \hdashline \addlinespace
\textbf{TA (BERT)}            & 108M  & \textbf{0.2440}      & \underline{9}  & \textbf{ 0.9267} & \underline{1.1707}  \\
\bottomrule
\end{tabular}
\end{subtable}
\end{table*}
\section{Ablation}
\label{app:ablation}
Based on the token importance predicted by the TA, various design choices are possible. We explore and compare the effectiveness of these variations. This study enhances our understanding of how generative LM interacts with token weights when learning data. The description of the components and experimental results are as follows.

\textbf{(1) Token-importance weight (ours) :} The original variation that utilizes token-importance weight predicted by TA for target-weighted learning. \textbf{(2) Known token masking :} Masking out the tokens in real-time when prediction and label matches. This method is intended to enhance “model awareness” in TA, as TA is more oriented to “task awareness.” \textbf{(3) Token weight dropping :} Among token-weight generated by TA, dropping weights that are below the top k\% levels. We tested 50\% and 80\%. Vanilla TA is the same as the threshold of 0\%. This method is intended to cut out noisy targets, as TA is supposed to assign lower weight to un-useful tokens. 

 \paragraph{Results} Known token masking does not yield better results compared to TAALM w/ token-importance weight. We hypothesize that the effect of known masking is limited because task awareness is already achieved when the loss of learned tokens is reduced. Test results on the TAALM w/ token weight dropping show that as the threshold increases, the top accuracy decreases. This suggests that some useful targets are mixed in among the lower weights, and it helps the model learn better somehow. On the contrary, not-to-forget accuracy slightly improves as the threshold increases. This seems as the effect of 1) cutting out noisy targets and 2) trade-off for lower learning. However, Total Knowledge is best on the TAALM w/ token-importance weight (ours). Overall experimental results indicate that, since TA is optimized to maximize task performance, adding heuristic interventions appears to produce suboptimal outcomes.

\begin{table}[h!]
\caption{\textsc{LAMA-ckl} performance of Llama2-7B based baselines.}

\hfill

\label{tab:ablation}

\hfill

\centering
\begin{tabular}{
 lllll
}
\toprule
                          & Top Acc      & Epoch     & NF Acc &Total Knowledge   \\ \hline
Finetune         & 0.1150        & 16     & 0.8174   & 0.9324 \\
TAALM w/ token-importance weight (\textbf{ours})          & \textbf{0.4290}     & \textbf{4} & 0.8983 & \textbf{1.3273} \\
TAALM w/ known token masking               & 0.3920   & \textbf{4} & 0.9075 & 1.2995 \\
TAALM w/ token weight dropping < 0.5       & \underline{0.4100}     & 7   & \underline{0.9148} & \underline{1.3248}   \\
TAALM w/ token weight dropping < 0.8       & 0.3850     & \textbf{4}   & \textbf{0.9267} & 1.3117  \\
\bottomrule
\end{tabular}

\end{table}

\begin{figure}[h!]
  \centerline{
 \begin{subfigure}{0.5\textwidth} 
    \centering
    \includegraphics[width=\linewidth]{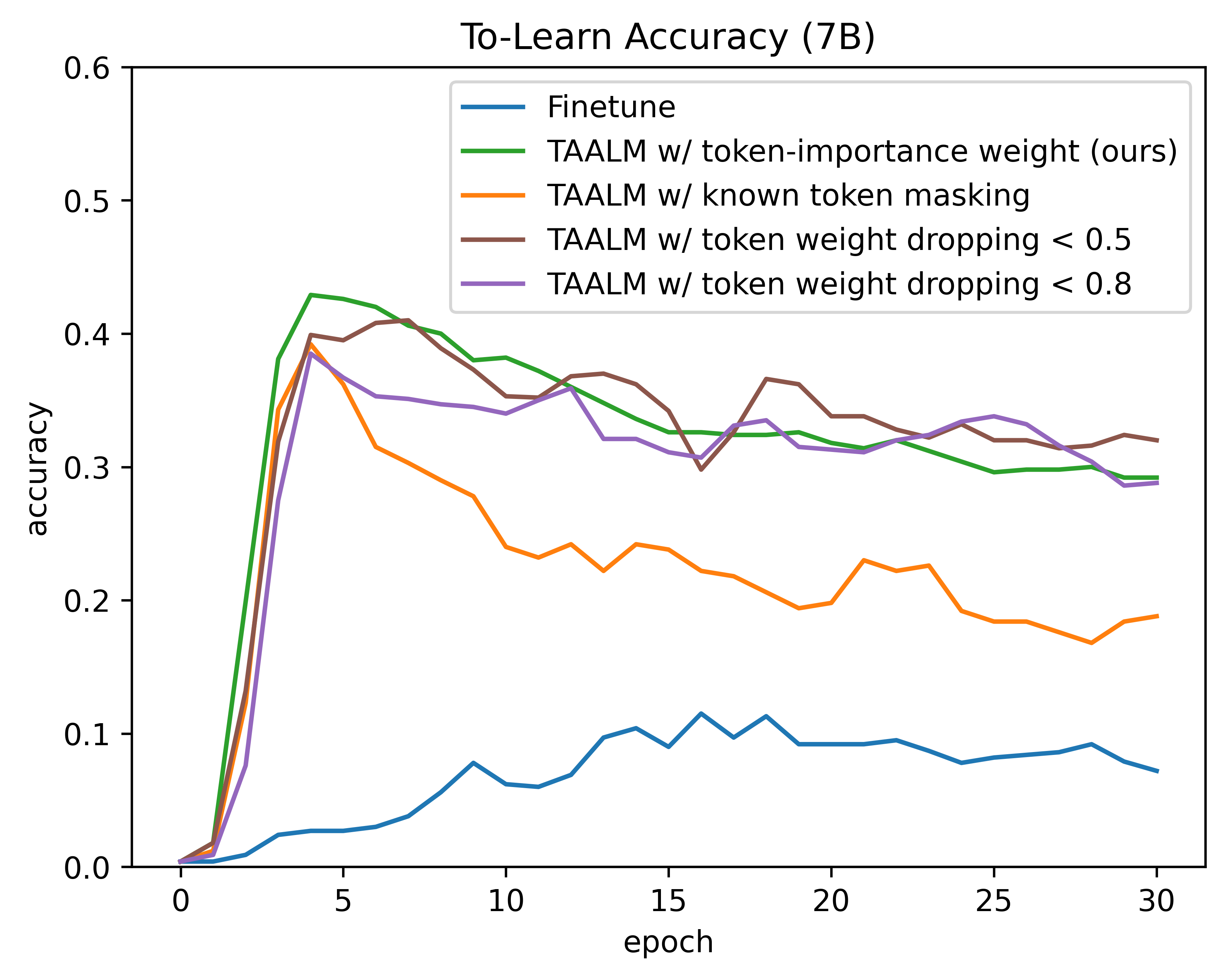}
    \end{subfigure}%
 \begin{subfigure}{0.5\textwidth} 
    \centering
    \includegraphics[width=\linewidth]{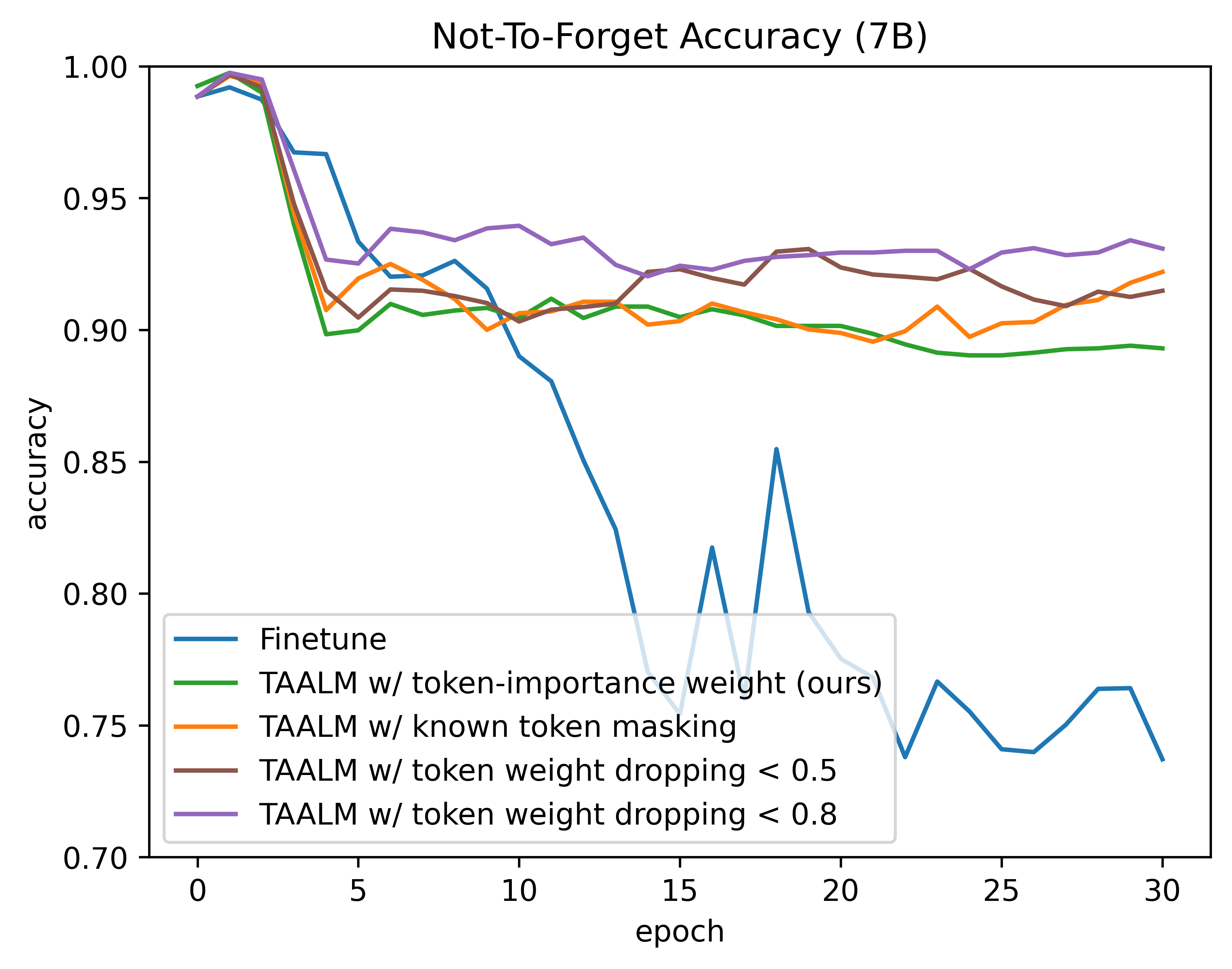}
    \end{subfigure}%
    }

  \caption{Ablation study on various design choices with token importance from Train-Attention (TA). TA is Tinyllama-1B, and the generation model is Llama2-7B.}
  
  \label{fig:oracle}
\end{figure}

\newpage
\section{Analysis on the attention pattern of TA}
\begin{figure}[h!]
\begin{subfigure}{0.99\textwidth}
  \includegraphics[width=0.99\textwidth]{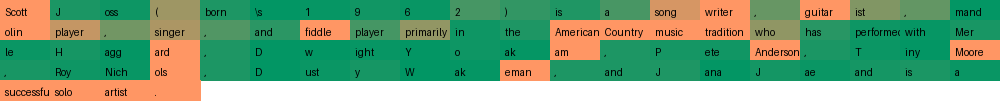}
  \caption{Heat map of token weights from Train-Attention trained on \textsc{LAMA-ckl}.}
  \label{fig:heatmap1}
\end{subfigure}

\begin{subfigure}{0.99\textwidth}
  \includegraphics[width=0.99\textwidth]{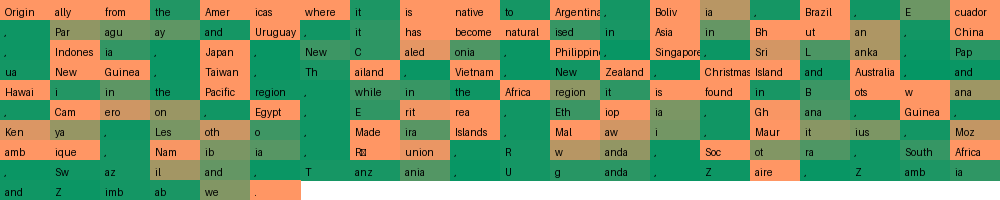}
  \caption{Heat map of token weights from Train-Attention trained on \textsc{LAMA-ckl}.}
  \label{fig:heatmap2}
\end{subfigure}

\begin{subfigure}{0.99\textwidth}
  \includegraphics[width=0.99\textwidth]{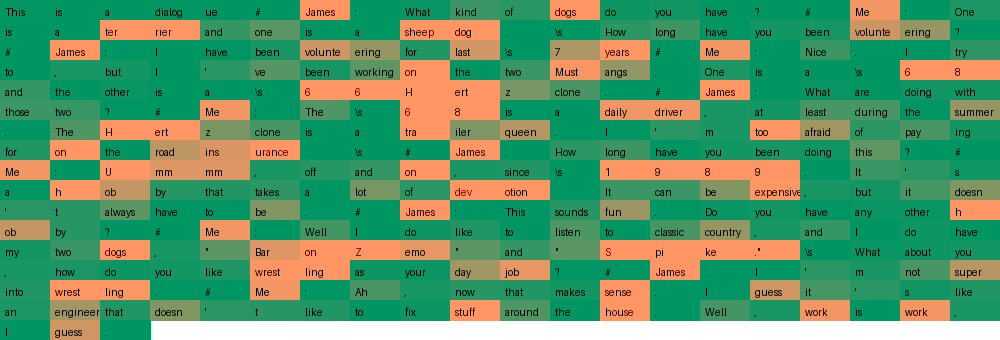}
  \caption{Heat map of token weights from Train-Attention trained on Multi-Session Chat dataset.}
  \label{fig:heatmap3}
\end{subfigure}
\caption{Heat map of token weights from Train-Attention. {\color{orange}Orange} color indicates higher weights.}
\end{figure}

TA is observed to generally assign attention to proper nouns, nouns, and verbs that contain the subject's character. The focus of attention seems to be diverse depending on the content of the text. For autobiographical texts, TA shows a tendency to focus on words that represent the person's occupation or major events (\autoref{fig:heatmap1}). In passages listing regional relations, TA pinpoints the names of locations (\autoref{fig:heatmap2}). This appears to be due to the consideration of probable queries. While TA (trained on \textsc{LAMA-ckl}) omits some words in the documents, it tends to not miss location names. This is likely because many queries in the \textsc{LAMA-ckl} benchmark involve location-related aspects (e.g., birthplaces, location of the workplace).

We also provide an attention map of TA trained on the multi-session chat (\autoref{fig:heatmap3}). Here, we regard prior dialogue sessions as data ($\mathcal{D}$) and an understanding of the next session as task ($\mathcal{T_D}$). Unlike Wikipedia documents, chit-chat dialogues contain fewer useful words, highlighting the necessity for TA. TA focuses on the interlocutor’s information like the occupation and pet's name.

\newpage

\section*{NeurIPS Paper Checklist}

\begin{enumerate}

\item {\bf Claims}
    \item[] Question: Do the main claims made in the abstract and introduction accurately reflect the paper's contributions and scope?
    \item[] Answer: \answerYes{} 
    \item[] Justification: The paper includes our mathematical formulation and quantitative experimental
results that reflect and justify the claims in our abstract and introduction.

\item {\bf Limitations}
    \item[] Question: Does the paper discuss the limitations of the work performed by the authors?
    \item[] Answer: \answerYes{} 
    \item[] Justification: The limitation section contains a discussion of our method's limitations.
    \item[] Guidelines:

\item {\bf Theory Assumptions and Proofs}
    \item[] Question: For each theoretical result, does the paper provide the full set of assumptions and a complete (and correct) proof?
    \item[] Answer: \answerNA{} 
    \item[] Justification: The paper does not include theoretical results.

    \item {\bf Experimental Result Reproducibility}
    \item[] Question: Does the paper fully disclose all the information needed to reproduce the main experimental results of the paper to the extent that it affects the main claims and/or conclusions of the paper (regardless of whether the code and data are provided or not)?
    \item[] Answer: \answerYes{} 
    \item[] Justification: We show how our meta-learning algorithm can be applied to the CKL benchmarks. We clearly explain our algorithm, and architecture with visual aids. We provide anonymized code for our quantitative experiments alongside clear instructions (README.md) for training and evaluation.

\item {\bf Open access to data and code}
    \item[] Question: Does the paper provide open access to the data and code, with sufficient instructions to faithfully reproduce the main experimental results, as described in supplemental material?
    \item[] Answer: \answerYes{} 
    \item[] Justification: We provide anonymized code for our quantitative experiments alongside clear instructions (README.md) for training and evaluation.

\item {\bf Experimental Setting/Details}
    \item[] Question: Does the paper specify all the training and test details (e.g., data splits, hyperparameters, how they were chosen, type of optimizer, etc.) necessary to understand the results?
    \item[] Answer: \answerYes{} 
    \item[] Justification:  Full training and testing details are in appendix. Full implementations of generative models and classifiers are included in code.

\item {\bf Experiment Statistical Significance}
    \item[] Question: Does the paper report error bars suitably and correctly defined or other appropriate information about the statistical significance of the experiments?
    \item[] Answer: \answerYes{} 
    \item[] Justification: For results on the main experiment (\textsc{LAMA-ckl} benchmark, Llama2-7B baselines), we present error ranges at ±2$\sigma$ derived from the number of trials (5), on Appendix~\ref{app:error_bar}.
    
\item {\bf Experiments Compute Resources}
    \item[] Question: For each experiment, does the paper provide sufficient information on the computer resources (type of compute workers, memory, time of execution) needed to reproduce the experiments?
    \item[] Answer: \answerYes{} 
    \item[] Justification: We include the name of the GPU we used in addition to the time of execution.

\item {\bf Code Of Ethics}
    \item[] Question: Does the research conducted in the paper conform, in every respect, with the NeurIPS Code of Ethics \url{https://neurips.cc/public/EthicsGuidelines}?
    \item[] Answer: \answerYes{} 
    \item[] Justification: We do not work with human participants, and all relevant datasets have been checked for privacy compliance prior to experiments and submission. We do not release any model that could be considered high-risk, and we offer a discussion of broader societal impacts in our discussion section.

\item {\bf Broader Impacts}
    \item[] Question: Does the paper discuss both potential positive societal impacts and negative societal impacts of the work performed?
    \item[] Answer: \answerYes{} 
    \item[] Justification: We discuss the potential positive and negative impacts of this work in \autoref{sec:conclusion}.
    
\item {\bf Safeguards}
    \item[] Question: Does the paper describe safeguards that have been put in place for responsible release of data or models that have a high risk for misuse (e.g., pretrained language models, image generators, or scraped datasets)?
    \item[] Answer: \answerNA{} 
    \item[] Justification: Our contribution does not include new datasets or pre-trained models that pose a risk of misuse.
    
\item {\bf Licenses for existing assets}
    \item[] Question: Are the creators or original owners of assets (e.g., code, data, models), used in the paper, properly credited and are the license and terms of use explicitly mentioned and properly respected?
    \item[] Answer: \answerYes{} 
    \item[] Justification: Code that we derive from earlier work is properly licensed and referenced.

\item {\bf New Assets}
    \item[] Question: Are new assets introduced in the paper well documented and is the documentation provided alongside the assets?
    \item[] Answer: \answerYes{} 
    \item[] Justification: We provide anonymized code for our quantitative experiments alongside clear instructions for training and evaluation.

\item {\bf Crowdsourcing and Research with Human Subjects}
    \item[] Question: For crowdsourcing experiments and research with human subjects, does the paper include the full text of instructions given to participants and screenshots, if applicable, as well as details about compensation (if any)? 
    \item[] Answer: \answerNA{} 
    \item[] Justification: No human subjects or crowdsourcing were involved in this research.

\item {\bf Institutional Review Board (IRB) Approvals or Equivalent for Research with Human Subjects}
    \item[] Question: Does the paper describe potential risks incurred by study participants, whether such risks were disclosed to the subjects, and whether Institutional Review Board (IRB) approvals (or an equivalent approval/review based on the requirements of your country or institution) were obtained?
    \item[] Answer: \answerNA{} 
    \item[] Justification: No human subjects were involved in this research.

\end{enumerate}

\end{document}